\documentclass{article}

\usepackage{jmlr2e}
\usepackage{paralist,url}
\usepackage[ruled]{algorithm2e}
\usepackage{microtype,amssymb,enumitem,multirow,algorithmic,mathtools,amsmath,bbold}

\SetKwComment{Comment}{$\triangleright$\ }{}

\newtheorem{assumption}{Assumption}

\newcommand{\bw}{{\boldsymbol w}}
\newcommand{\bx}{{\boldsymbol x}}
\newcommand{\bs}{{\boldsymbol s}}
\newcommand{\by}{{\boldsymbol y}}
\newcommand{\bz}{{\boldsymbol z}}

\newcommand{\bd}{{\boldsymbol d}}
\newcommand{\AL}{{\boldsymbol \alpha}}

\newcommand{\M}{\mathcal{M}}
\newcommand{\cP}{\mathcal{I}}
\newcommand{\cPX}{\mathcal{I}^X}
\newcommand{\cPg}{\mathcal{I}^g}

\newcommand{\R}{\mathbb{R}}

\newcommand{\trace}{\mbox{\rm trace}}
\newcommand{\diag}{\text{diag}}

\newcommand{\TheTitle}{A Distributed Quasi-Newton Algorithm for
Primal and Dual Regularized Empirical Risk Minimization}

\begin{document}
\title{{\TheTitle}}
\author{\name{Ching-pei Lee}
\email{leechingpei@gmail.com}\\
	\addr Department of Mathematics\\
	\addr National University of Singapore\\
	\addr Singapore 119076\\
\name{Cong Han Lim}
\email{clim9@wisc.edu}\\
	\addr Wisconsin Institute for Discovery\\
	\addr University of Wisconsin-Madison\\
	\addr Madison, Wisconsin 53715\\
\name{Stephen~J. Wright}
\email{swright@cs.wisc.edu}\\
\addr Department of Computer Sciences\\
	\addr University of Wisconsin-Madison\\
	\addr Madison, Wisconsin 53706
}
\editor{}

\maketitle
\begin{abstract}
We propose a communication- and computation-efficient distributed
optimization algorithm using second-order information for solving
empirical risk minimization (ERM) problems with a nonsmooth
regularization term.
Our algorithm is applicable to both the primal and the dual ERM problem.
Current second-order and quasi-Newton methods for this
problem either do not work well in the distributed setting or
work only for specific regularizers. Our algorithm uses successive
quadratic approximations of the smooth part, and we describe how to
maintain an approximation of the (generalized) Hessian and solve
subproblems efficiently in a distributed manner.
When applied to the distributed dual ERM problem, unlike state of the
art that takes only the block-diagonal part of the Hessian, our
approach is able to utilize global curvature information and is thus
magnitudes faster.
The proposed method enjoys global linear convergence for
a broad range of non-strongly convex problems that includes the most
commonly used ERMs, thus requiring lower communication complexity.
It also converges on non-convex problems, so has the
potential to be used on applications such as deep learning.
Computational results demonstrate that our method significantly
improves on communication cost and running time over the current
state-of-the-art methods.

\end{abstract}


\section{Introduction}
\label{sec:intro}
We consider using multiple machines to solve the following regularized
problem
\begin{equation}
\min_{\bw \in \R^d} \quad P\left (\bw \right)  \coloneqq
	\xi\left( X^\top \bw \right) + g\left( \bw \right),
\label{eq:primal}
\end{equation}
where $X$ is a $d$ by $n$ real-valued matrix, and $g$ is a convex,
closed, and extended-valued proper function that can be
nondifferentiable, or its dual problem
\begin{equation}
\min_{\AL \in \R^n} \quad D\left(\AL\right)  \coloneqq g^*\left(X\AL
\right) +
	\xi^*\left(-\AL \right),
\label{eq:dual}
\end{equation}
where for any given function $f(\cdot)$, $f^*$ denotes its convex
conjugate
\begin{equation*}
	f^*\left(z\right) \coloneqq \max_{y}\quad z^\top y - f\left(y\right).
\end{equation*}
Each column of $X$ represents a single data point or instance, and we
assume that the set of data points is partitioned and spread across
$K > 1$ machines (i.e.\ distributed \emph{instance-wise}). We write
$X$ as
\begin{equation}
	X \coloneqq \left[X_1, X_2,\dotsc, X_K \right]
\label{eq:storage}
\end{equation}
where $X_k$ is stored exclusively on the $k$th machine. The dual
variable $\AL$ is formed by concatenating $\AL_1,\AL_2,\dotsc,\AL_K$
where $\AL_k$ is the dual variable corresponding to $X_k$. We let
$\cPX_1,\dotsc,\cPX_K$ denote the indices of the columns of $X$
corresponding to each of the $X_k$ matrices. We further assume that
$\xi$ shares the same block-separable structure and can be written as
follows:
\begin{equation}
\xi\left( X^\top \bw \right) = \sum_{k=1}^K \xi_k\left(X_k^\top \bw\right),
\label{eq:finitesum}
\end{equation}
and therefore in \eqref{eq:dual}, we have
\begin{equation}
	\xi^*\left( -\AL \right) = \sum_{k=1}^K \xi^*_k\left( -\AL_k
	\right).
	\label{eq:finitesumdual}
\end{equation}
For the ease of description and unification, when solving the primal
problem, we also assume that there exists some partition
$\cPg_1,\dotsc,\cPg_K$ of $\{1,\dotsc,d\}$
and $g$ is block-separable according to the partition:
\begin{equation}
	g\left(\bw\right) = \sum_{k=1}^K g_k\left(\bw_{\cPg_k}\right),
	\label{eq:blockseparable}
\end{equation}
though our algorithm can be adapted for non-separable $g$ with minimal
modification, see the preliminary version \cite{LeeLW18a}.

When we solve the primal problem \eqref{eq:primal}, $\xi$ is assumed
to be a differentiable function with Lipschitz continuous gradients,
and is allowed to be nonconvex.
On the other hand, when the dual problem \eqref{eq:dual} is
considered, for recovering the primal solution,
we require strong convexity on $g$ and convexity on $\xi$,
and $\xi$ can be either nonsmooth but Lipschitz continuous (within the
area of interest), or Lipschitz continuously differentiable.
Note that strong convexity of $g$ implies that $g^*$ is
Lipschitz-continuously differentiable \citep[Part E, Theorem 4.2.1 and
Theorem 4.2.2]{HirL01a}, making \eqref{eq:dual} have the same
structure as \eqref{eq:primal} such that both problems have one smooth
and one nonsmooth term.
There are several reasons for considering the alternative dual problem.
First, when $\xi$ is nonsmooth, the primal problem becomes hard to
solve as both terms are nonsmooth, meanwhile in the dual problem,
$\xi^*$ is guaranteed to be smooth.
Second, the number of variables in the primal and the dual problem are
different.
In our algorithm whose spatial and temporal costs are positively
correlated to the number of variables, when the data set has much
higher feature dimension than the number of data points, solving the
dual problem can be more economical.


The bottleneck in performing distributed optimization is often the
high cost of communication between machines. For \eqref{eq:primal} or
\eqref{eq:dual},
the time required to retrieve $X_k$ over a network can greatly exceed
the time needed to compute $\xi_k$ or its gradient with locally stored
$X_k$.  Moreover, we incur a delay at the beginning of each round of
communication due to the overhead of establishing connections between
machines. This latency prevents many efficient single-core algorithms
such as coordinate descent (CD) and stochastic gradient and their
asynchronous parallel variants from being employed in large-scale
distributed computing setups. Thus, a key aim of algorithm design for
distributed optimization is to improve the communication efficiency
while keeping the computational cost affordable. Batch methods are
preferred in this context, because fewer rounds of communication occur
in distributed batch methods.




When the objective is smooth, many batch methods can be used
directly in distributed environments to optimize them.
For example, Nesterov's accelerated gradient (AG) \citep{Nes83a} enjoys
low iteration complexity, and since each iteration of AG only requires
one round of communication to compute the new gradient, it also has
good communication complexity.
Although its supporting theory is not particularly strong, the
limited-memory BFGS (LBFGS) method \citep{LiuN89a} is popular among
practitioners of distributed optimization.
It is the default algorithm for solving $\ell_2$-regularized smooth ERM
problems in Apache Spark's distributed machine learning library
\citep{Men16a}, as it is empirically much faster than AG (see, for
example, the experiments in \citet{WanLL16a}). Other batch methods
that utilize the Hessian of the objective in various ways are also
communication-efficient under their own additional assumptions
\citep{ShaSZ14a, ZhaL15a, LeeWCL17a,ZhuCJL15a,LinTLL14a}.

However, when the objective is nondifferentiable, neither LBFGS nor Newton's
method can be applied directly.  Leveraging curvature information from
the smooth part ($\xi$ in the primal or $g^*$ in the dual)
can still be beneficial in this setting. For example, the orthant-wise
quasi-Newton method OWLQN \citep{AndG07a} adapts the LBFGS algorithm
to the special nonsmooth case in which $g(\cdot) \equiv \|\cdot\|_1$
for \eqref{eq:primal}, and is popular for distributed optimization of
$\ell_1$-regularized ERM problems. Unfortunately, extension of this
approach to other nonsmooth $g$ is not well understood, and the
convergence guarantees are only asymptotic, rather than global.
Another example is that for \eqref{eq:dual}, state of the art
distributed algorithms \citep{Yan13a,LeeC17a,ZheXXZ17a} utilize
block-diagonal entries of the real Hessian of $g^*(X\AL)$.
%

To the best of our knowledge, for ERMs with \emph{general} nonsmooth
regularizers in the instance-wise storage setting,
proximal-gradient-like methods \citep{WriNF09a,BecT09a,Nes13a} are the
only practical distributed optimization algorithms with convergence
guarantees for the primal problem \eqref{eq:primal}. Since these
methods barely use the curvature information of the smooth part (if at
all), we suspect that proper utilization of second-order information
has the potential to improve convergence speed and therefore
communication efficiency dramatically. As for algorithms solving the
dual problem \eqref{eq:dual}, computing $X\AL$ in the instance-wise
storage setting requires communicating a $d$-dimensional vector, and
only the block-diagonal part of $\partial^2_{\AL} g^*(X\AL)$ can be
obtained easily. Therefore, global curvature information is not
utilized in existing algorithms, and we expect that utilizing global
second-order information of $g^*$ can also provide substantial
benefits over the block-diagonal approximation approaches.
We thus propose a practical distributed inexact variable-metric
algorithm that can be applied to both \eqref{eq:primal} and
\eqref{eq:dual}.
Our algorithm uses gradients and updates information from previous
iterations to estimate curvature of the smooth part in a
communication-efficient manner. We describe construction of this
estimate and solution of the corresponding subproblem. We also provide
convergence rate guarantees, which also bound communication
complexity.  These rates improve on existing distributed methods, even
those tailor-made for specific regularizers.

More specifically, We propose a distributed inexact
proximal-quasi-Newton-like algorithm that can be used to solve both
\eqref{eq:primal} and \eqref{eq:dual} under the instance-wise split
setting that share the common structure of having a smooth term $f$
and a nonsmooth term $\Psi$.
At each iteration with the current iterate $x$, our algorithm utilizes
the previous update directions and gradients to construct a
second-order approximation of the smooth part $f$ by the LBFGS method,
and approximately minimizes this quadratic term plus the nonsmooth
term $\Psi$ to obtain an update iteration $p$.
\begin{equation}
	p \approx \arg\min_{p}\, Q_H(p;x),
	\label{eq:quadratic}
\end{equation}
where $H$ is the LBFGS approximation of the Hessian of $f$ at $x$,
and
\begin{equation}
Q_H(p;x)\coloneqq \nabla f(x)^\top p + \frac12 p^\top H p +
	\Psi(x+p) - \Psi(x).
\label{eq:Q}
\end{equation}

For the primal problem \eqref{eq:primal}, we believe that this work is
the first to propose, analyze, and implement a practically feasible
distributed optimization method for solving \eqref{eq:primal} with
general nonsmooth regularizer $g$ under the instance-wise storage
setting.
For the dual problem \eqref{eq:dual}, our algorithm is the first to
suggest an approach that utilizes global curvature information under
the constraint of distributed data storage.
This usage of non-local curvature information greatly improves upon
state of the art for the distributed dual ERM problem which uses the
block-diagonal parts of the Hessian only.
An obvious drawback of the block-diagonal approach is that the
convergence deteriorates with the number of machines, as more and more
off-block-diagonal entries are ignored.
In the extreme case, where there are $n$ machines such that each
machine stores only one column of $X$, the block-diagonal approach
reduces to a scaled proximal-gradient algorithm and the convergence
is expected to be extremely slow.
On the other hand, our algorithm has convergence behavior independent
of number of machines and data distribution over nodes, and is thus
favorable when many machines are used.
Our approach has both good communication and computational
complexities, unlike certain approaches that focus only on
communication at the expense of computation (and ultimately overall
time).
\subsection{Contributions}
We summarize our main contributions as follows.
\begin{itemize}
\item The proposed method is the first real distributed second-order
	method for the dual ERM problem that utilizes global curvature
	information of the smooth part.
	Existing second-order methods use only the block-diagonal part of
	the Hessian and suffers from asymptotic convergence speed as slow
	as proximal gradient, while our method enjoys fast convergence
	throughout.
	Numerical results show that our inexact proximal-quasi-Newton
	method is magnitudes faster than state of the art for distributed
	optimizing the dual ERM problem.
\item We propose the first distributed algorithm for primal ERMs with
	general nonsmooth regularizers \eqref{eq:primal} under the
	instance-wise split setting.
	Prior to our work, existing algorithms are either for a specific
	regularizer (in particular the $\ell_1$ norm) or for the
	feature-wise split setting, which is often impractical.
	In particular, it is usually easier to generate new data
	points than to generate new features, and each time new data points
	are obtained from one location, one needs to distribute their entries
	to different machines under the feature-wise setting.
\item The proposed framework is applicable to both primal and dual ERM
	problems under the same instance-wise split setting, and the
	convergence speed is not deteriorated by the number of machines.
	Existing methods that applicable to both problems can deal with
	feature-wise split for the primal problem only,
	and their convergence degrades with the number of machines used,
	and are thus not suitable for large-scale applications where
	thousands of or more machines are used.
	This unification also reduces two problems into one and facilitates
	future development for them.
\item Our analysis provides sharper convergence guarantees and
	therefore better communication efficiency.  In particular, global
	linear convergence for a broad class of non-strongly convex
	problems that includes many popular ERM problems are shown, and an
	early linear convergence to rapidly reach a medium solution
	accuracy is proven for convex problems.
\end{itemize}

\subsection{Organization}
We first describe the general distributed algorithm in
Section~\ref{sec:alg}.  Convergence guarantee, communication
complexity, and the effect of the subproblem solution inexactness are
analyzed in Section~\ref{sec:analysis}.
Specific details for applying our algorithm respectively on the primal
and the dual problem are given in Section \ref{sec:primaldual}.
Section~\ref{sec:related} discusses related works, and empirical
comparisons are conducted in Section~\ref{sec:exp}.
Concluding observations appear in Section~\ref{sec:conclusions}.

\subsection{Notation and Assumptions}
We use the following notation.
\begin{itemize}
\item $\|\cdot\|$ denotes the 2-norm, both for vectors and for matrices.
\item Given any symmetric positive semi-definite matrix $H \in
	\R^{d\times d}$ and any vector $p \in \R^d$,
	$\|p\|_H$ denotes the semi-norm $\sqrt{p^\top H p}$.
\end{itemize}

In addition to the structural assumptions of distributed instance-wise
storage of $X$ in \eqref{eq:storage} and the block separability of
$\xi$ in \eqref{eq:finitesum}, we also use the following assumptions
throughout this work.  When we solve the primal problem, we assume the
following.
\begin{assumption}
\label{assum:primal}
The regularization term $g(\bw)$ is convex, extended-valued, proper,
and closed.
The loss function $\xi(X^\top \bw)$ is $L$-Lipschitz continuously
differentiable with respect to $\bw$ for some $L>0$.
That is,
\begin{equation}
	\left\|X^\top \xi'\left( X^\top \bw_1 \right) - X^\top \xi' \left( X^\top
	\bw_2 \right)\right\| \leq L \left\|\bw_1 -
	\bw_2\right\|,
	\forall \bw_1, \bw_2 \in \R^d.
	\label{eq:Lipschitz}
\end{equation}
\end{assumption}

On the other hand, when we consider solving the dual problem, the
following is assumed.
\begin{assumption}
\label{assum:dual}
Both $g$ and $\xi$ are convex.
$g^*(X\AL)$ is $L$-Lipschitz continuously differentiable with respect
to $\AL$.
Either $\xi^*$ is $\sigma$-strongly convex for some $\sigma > 0$, or
the loss term $\xi(X^\top \bw)$ is $\rho$-Lipschitz continuous for some
$\rho$.
\end{assumption}
Because a function is $\rho$-Lipschitz continuously
differentiable if and only if its conjugate is $(1/\rho)$-strongly
convex \citep[Part E, Theorem 4.2.1 and Theorem 4.2.2]{HirL01a},
Assumption \ref{assum:dual} implies that $g$ is $\|X^\top X\|/L$-strongly
convex.
From the same reasoning, $\xi^*$ is $\sigma$-strongly convex if only
if $\xi$ is $(1 / \sigma)$ Lipschitz continuously differentiable.
Convexity of the primal problem in Assumption \ref{assum:dual}
together with Slater's condition guarantee strong duality
\citet[Section 5.2.3]{BoyV04a}, which then ensures \eqref{eq:dual} is
indeed an alternative to \eqref{eq:primal}.
Moreover, from KKT conditions, any optimal solution $\AL^*$ for
\eqref{eq:dual} gives us a primal optimal solution $\bw^*$ for
\eqref{eq:primal} through
\begin{equation}
	\bw^* = \nabla g^*(X\AL^*).
\label{eq:wopt}
\end{equation}

\section{Algorithm}
\label{sec:alg}

We describe and analyze a general algorithmic scheme that can be
applied to solve both the primal \eqref{eq:primal} and dual
\eqref{eq:dual} problems under the instance-wise distributed data
storage scenario \eqref{eq:storage}. In Section \ref{sec:primaldual},
we discuss how to efficiently implement particular steps of this scheme
for \eqref{eq:primal} and \eqref{eq:dual}.

Consider a general problem of the form
\begin{equation}
	\min_{x\in \R^N}\quad F(x) \coloneqq f(x) + \Psi(x),
	\label{eq:f}
\end{equation}
where $f$ is $L$-Lipschitz continuously differentiable for some $L >
0$ and $\Psi$ is convex, closed, proper, extended valued, and
block-separable into $K$ blocks. More specifically, we can write
$\Psi(x)$ as
\begin{equation}
	\Psi(x) = \sum_{k=1}^K \Psi_k(x_{\cP_k}).
	\label{eq:psiblockseparable}
\end{equation}
where $\cP_1,\dotsc,\cP_K$ partitions $\{1,\dotsc,N\}$.

We assume as well that for the $k$th machine, $\nabla_{\cP_k} f(x)$ can be
obtained easily after communicating a vector of size $O(d)$ across
machines, and postpone the detailed gradient calculation until we
discuss specific problem structures in later sections.
Note that this $d$ is the primal variable dimension in
\eqref{eq:primal} and is independent of $N$.

The primal and dual problems are specific cases of the general
form \eqref{eq:f}. For the primal problem
\eqref{eq:primal} we let $N =d$, $x = \bw$, $f(\cdot) = \xi(X^\top
\cdot)$, and $\Psi(\cdot) = g(\cdot)$. The block-separability of
$g$ \eqref{eq:blockseparable} gives the desired block-separability of
$\Psi$ \eqref{eq:psiblockseparable}, and the Lipschitz-continuous
differentiability of $f$ comes from Assumption \ref{assum:primal}.
For the dual problem \eqref{eq:dual}, we have $N = n$, $x = \AL$,
$f(\cdot) = g^*(X\cdot)$, and $\Psi(\cdot) = \xi^*(-\cdot)$. The
separability follows from \eqref{eq:finitesumdual}, where the partition
\eqref{eq:psiblockseparable} reflects the data partition in
\eqref{eq:storage} and
Lipschitz continuity from Assumption \ref{assum:dual}.

Each iteration of our algorithm has four main steps -- (1) computing
the gradient $\nabla f(x)$, (2) constructing an approximate Hessian
$H$ of $f$, (3) solving a quadratic approximation subproblem to find
an update direction $p$, and finally (4) taking a step $x + \lambda p$
either via line search or trust-region approach. The gradient
computation step and part of the line search process is dependent on
whether we are solving the primal or dual problem, and we defer the
details to Section \ref{sec:primaldual}. The approximate Hessian $H$
comes from the LBFGS algorithm \cite{LiuN89a}. To compute the update
direction, we approximately solve \eqref{eq:quadratic},
where $Q_H$ consists of a quadratic approximation to $f$ and the
regularizer $\Psi$ as defined in \eqref{eq:Q}.
We then use either a line search procedure to determine a suitable
stepsize $\lambda$ and perform the update $x \leftarrow x + \lambda
p$, or use some trust-region-like techniques to decide whether to
accept the update direction with unit step size.

We now discuss the following issues in the distributed setting:
communication cost in distributed environments, the choice and
construction of $H$ that have low cost in terms of both communication
and per machine computation, procedures for solving
\eqref{eq:quadratic}, and the line search and trust-region procedures
for ensuring sufficient objective decrease.

\subsection{Communication Cost Model}
For the ease of description, we assume the {\em allreduce} model of
MPI \citep{MPI94a} throughout the work, but it is also straightforward
to extend the framework to a master-worker platform.
Under this {\em allreduce} model, all machines simultaneously fulfill
master and worker roles, and for any distributed operations that
aggregate results from machines, the resultant is broadcast to all
machines.

This can be considered as equivalent to conducting one map-reduce
operation and then broadcasting the result to all nodes.  The
communication cost for the allreduce operation on a $d$-dimensional
vector under this model is
\begin{equation}
	\log \left( K \right) T_{\text{initial}} + d T_{\text{byte}},
	\label{eq:commcost}
\end{equation}
where $T_{\text{initial}}$ is the latency to establish connection
between machines, and $T_{\text{byte}}$ is the per byte transmission
time (see, for example, \citet[Section 6.3]{ChaHPV07a}).

The first term in \eqref{eq:commcost} also explains why batch methods
are preferable. Even if methods that frequently update the iterates
communicate the same amount of bytes, it takes more rounds of
communication to transmit the information, and the overhead of
$\log (K) T_{\text{initial}}$ incurred at every round of communication
makes this cost dominant, especially when $K$ is large.

In subsequent discussion, when an allreduce operation is performed on
a vector of dimension $O(d)$, we simply say that a round of $O(d)$
communication is conducted. We omit the latency term since batch
methods like ours tend to have only a small constant number of rounds
of communication per iteration. By contrast, non-batch methods such as
CD or stochastic gradient require number of communication rounds per
epoch equal to data size or dimension, and therefore face much more
significant latency issues.

\subsection{Constructing a good $H$ efficiently}
\label{subsec:products}


We use the Hessian approximation constructed by the LBFGS algorithm
\citep{LiuN89a} as our $H$ in \eqref{eq:Q}, and propose a way to
maintain it efficiently in a distributed setting.
In particular, we show that most vectors involved can be stored
perfectly in a distributed manner in accord with the partition $\cP_k$
in \eqref{eq:psiblockseparable}, and this distributed storage further facilitates
parallelization of most computation.
Note that the LBFGS algorithm works even if the smooth part is not
twice-differentiable, see Lemma \ref{lemma:sparsa}.
In fact, Lipschitz continuity of the gradient implies that the
function is twice-differentiable almost everywhere, and generalized
Hessian can be used at the points where the smooth part is not
twice-differentiable. In this case, the LBFGS approximation is for the
generalized Hessian.

Using the compact representation in \citet{ByrNS94a}, given a
prespecified integer $m > 0$, at the $t$th iteration for $t > 0$, let
$m(t) \coloneqq \min(m,t)$, and define
\begin{equation*}
\bs_i \coloneqq x^{i+1} - x^i, \quad
\by_i \coloneqq \nabla f (x^{i+1}) - \nabla f(x^i),
\quad \forall i.
\end{equation*}
The LBFGS Hessian approximation matrix is
\begin{equation}
	H_t = \gamma_t I - U_t M_t^{-1} U_t^\top,
\label{eq:Hk}
\end{equation}
where
\begin{equation}
\label{eq:M}
U_t \coloneqq \left[\gamma_t S_t, Y_t\right],\quad
	M_t \coloneqq \left[\begin{array}{cc}
		\gamma_t S_t^\top S_t, & L_t\\
		L_t^\top & - D_t \end{array}\right],\quad
\gamma_t \coloneqq \frac{\by_{t-1}^\top \by_{t-1}}{\bs_{t-1}^\top \by_{t-1}},
\end{equation}
and
\begin{subequations} \label{eq:updates}
\begin{align}
S_t &\coloneqq \left[\bs_{t-m(t)}, \bs_{t-m(t)+1},\dotsc, \bs_{t-1}\right],\\
Y_t &\coloneqq \left[\by_{t-m(t)}, \by_{t-m(t)+1},\dotsc, \by_{t-1}\right],\\
D_t &\coloneqq \diag\left(\bs_{t-m(t)}^\top \by_{t-m(t)}, \dotsc,
	\bs_{t-1}^\top \by_{t-1}\right),\\
\left(L_t\right)_{i,j} &\coloneqq
	\begin{cases}
	\bs_{t - m(t) - 1 + i}^\top \by_{t - m(t) - 1 + j}, &\text{
		if } i > j,\\
	0, &\text{ otherwise.}
	\end{cases}
\end{align}
\end{subequations}
For $t=0$ where no $\bs_i$ and $\by_i$ are available, we
either set $H_0 \coloneqq a_0 I$ for some positive scalar $a_0$,
or use some Hessian approximation constructed using local data.
More details are given in Section \ref{sec:primaldual} when we
discuss the primal and dual problems individually.

If $f$ is not strongly convex, it is possible that \eqref{eq:Hk} is
only positive semi-definite, making the subproblem
\eqref{eq:quadratic} ill-conditioned.  In this case, we follow \citet{LiF01a},
taking the $m$ update pairs to be the most recent $m$ iterations for
which the inequality
\begin{equation}
\bs_i^\top \by_i \geq \delta \bs_i^\top \bs_i
\label{eq:safeguard}
\end{equation}
is satisfied, for some predefined $\delta > 0$. It can be shown that
this safeguard ensures that $H_t$ are always positive definite and the
eigenvalues are bounded within a positive range. For a proof in the
case that $f$ is twice-differentiable, see, for example, the appendix
of \citet{LeeW17a}. For completeness, we provide a proof without the
assumption of twice-differentiability of $f$ in Lemma
\ref{lemma:sparsa}.

To construct and utilize this $H_t$ efficiently, we store
$(U_t)_{\cP_k,:}$ on the $k$th machine, and all machines keep a copy of
the whole $M_t$ matrix as usually $m$ is small and this is affordable.
Under our assumption, on the $k$th machine, the local gradient
$\nabla_{\cP_k} f$ can be obtained, and we will show how to compute the
update direction $p_{\cP_k}$ locally in the next subsection.
Thus, since $\bs_i$ are just the update direction $p$ scaled by the step
size $\lambda$, it can be obtained without any additional communication.
All the information needed to construct $H_t$ is hence available locally on
each machine.

We now consider the costs associated with the matrix $M_t^{-1}$.
The matrix $M_t$, but not its inverse, is maintained for easier update.
In practice, $m$ is usually much smaller than $N$, so the $O(m^3)$
cost of inverting the matrix directly is insignificant compared to the
cost of the other steps.
On contrary, if $N$ is large, the computation of the inner products
$\bs_i^\top \by_j$ and $\bs_i^\top \bs_j$ can be the bottleneck in
constructing $M_t^{-1}$. We can significantly reduce this cost by
computing and maintaining the inner products in parallel and
assembling the results with $O(m)$ communication cost.
At the $t$th iteration, given the new $\bs_{t-1}$, because $U_t$ is
stored disjointly on the machines, we compute the inner products of
$\bs_{t-1}$ with both $S_t$ and $Y_t$ in parallel via the summations
\begin{equation*}
	\sum_{k=1}^K \left( (S_t)_{\cP_k,:}^\top (\bs_{t-1})_{\cP_k} \right),\quad
	\sum_{k=1}^K \left( (Y_t)_{\cP_k,:}^\top (\bs_{t-1})_{\cP_k} \right),
\end{equation*}
requiring $O(m)$ communication of the partial sums on each machine. We
keep these results until $\bs_{t-1}$ and $\by_{t-1}$ are discarded, so
that at each iteration, only $2m$ (not $O(m^2)$) inner products are
computed.


\subsection{Solving the Quadratic Approximation Subproblem to Find Update Direction}
\label{subsec:sparsa}
The matrix $H_t$ is generally not diagonal, so there is no easy
closed-form solution to \eqref{eq:quadratic}.  We will instead use
iterative algorithms to obtain an approximate solution to this
subproblem.
In single-core environments, coordinate descent (CD) is one of the
most efficient approaches for solving \eqref{eq:quadratic}
\citep{YuaHL12a, KaiYDR14a, SchT16a}.
When $N$ is not too large, instead of the distributed approach we
discussed in the previous section, it is possible to construct $H_t$
on all machines.
In this case, a local CD process can be applied on all machines to
save communication cost, in the price that all machines conduct
the same calculation and the additional computational power from
multiple machines is wasted.
The alternative approach of applying proximal-gradient
methods to \eqref{eq:quadratic} may be more efficient in
distributed settings, since they can be parallelized with
little communication cost for large $N$.


The fastest proximal-gradient-type methods are accelerated gradient
(AG) \citep{BecT09a,Nes13a} and SpaRSA \citep{WriNF09a}. SpaRSA is a basic
proximal-gradient method with spectral initialization of the parameter
in the prox term. SpaRSA has a few key advantages over AG despite
its weaker theoretical convergence rate guarantees. It tends to be
faster in the early iterations of the algorithm \citep{YanZ11a}, thus
possibly yielding a solution of acceptable accuracy in fewer
iterations than AG. It is also a descent method, reducing the
objective $Q_H$ at every iteration, which ensures that the solution
returned is at least as good as the original guess $p = 0$

In the rest of this subsection, we will describe a distributed
implementation of SpaRSA for \eqref{eq:quadratic}, with $H$ as
defined in \eqref{eq:Hk}. The major computation is obtaining the
gradient of the smooth (quadratic) part of \eqref{eq:Q},
and thus with minimal modification, AG can be used with the same per
iteration cost.  To distinguish between the iterations of our main
algorithm (i.e. the entire process required to update $x$ a single
time) and the iterations of SpaRSA, we will refer to them by
\emph{main iterations} and \emph{SpaRSA iterations} respectively.


Since $H$ and $x$ are fixed in this subsection, we will write
$Q_H(\cdot;x)$ simply as $Q(\cdot)$. We denote the $i$th iterate of
the SpaRSA algorithm as $p^{(i)}$, and we initialize $p^{(0)}
= 0$ whenever there is no obviously better choice.
We denote the smooth part of $Q_H$ by $\hat f(p)$, and the nonsmooth
$\Psi(x+p)$ by $\hat \Psi(p)$.
\begin{equation}
	\hat f(p) \coloneqq \nabla f(x)^\top p + \frac12 p^\top H p,\quad
	\hat \Psi(p) \coloneqq \Psi(x+p) - \Psi(x).
	\label{eq:quadfg}
\end{equation}
At the $i$th iteration of SpaRSA, we define
\begin{equation}
u^{(i)}_{\psi_i} \coloneqq p^{(i)} -  \frac{\nabla \hat
	f(p^{(i)})}{ \psi_i},
	\label{eq:udef}
\end{equation}
and solve the following subproblem:
\begin{equation}
p^{(i+1)} = \arg \min_{p} \, \frac{1}{2} \left\|p -
u^{(i)}_{\psi_i} \right\|^2 + \frac{\hat \Psi(p)}{\psi_i},
\label{eq:dk}
\end{equation}
where $\psi_i$ is defined by the following ``spectral'' formula:
\begin{equation}
	\psi_i = \frac{\left(p^{(i)} - p^{(i-1)}\right)^\top \left(\nabla
	\hat f(p^{(i)}) - \nabla \hat
	f(p^{(i-1)})\right)}{\left\|p^{(i)} - p^{(i-1)}\right\|^2}.
	\label{eq:psi}
\end{equation}
When $i=0$, we use a pre-assigned value for $\psi_0$ instead.  (In our
LBFGS choice for $H_t$, we use the value of $\gamma_t$ from
\eqref{eq:M} as the initial estimate of $\psi_0$.)  The exact minimizer of
\eqref{eq:dk} can be difficult to compute for general $\Psi$.
However, approximate solutions of \eqref{eq:dk} suffice to provide a
convergence rate guarantee for solving \eqref{eq:quadratic}
\citep{SchRB11a, SchT16a, GhaS16a, LeeW18a}.
Since it is known (see Lemma \ref{lemma:sparsa}) that the eigenvalues
of $H$ are upper- and lower-bounded in a positive range after the
safeguard \eqref{eq:safeguard} is applied, we can guarantee that this
initialization of $\psi_i$ is bounded within a positive range; see
Section~\ref{sec:analysis}.
The initial value of $\psi_i$ defined in \eqref{eq:psi} is increased
successively by a chosen constant factor $\beta>1$, and $p^{(i+1)}$
is recalculated from \eqref{eq:dk}, until the following sufficient
decrease criterion is satisfied:
\begin{equation}
	Q\left(p^{(i+1)}\right) \leq Q\left(p^{(i)}\right) -
	\frac{\sigma_0 \psi_i}{2} \left\|p^{(i+1)} -
	p^{(i)}\right\|^2,
	\label{eq:accept}
\end{equation}
for some specified $\sigma_0 \in (0,1)$.
Note that the evaluation of $Q(p)$ needed in \eqref{eq:accept} can
be done efficiently through a parallel computation of
\begin{equation*}
	\sum_{k=1}^K \frac12 \left(\nabla_{\cP_k} \hat f\left( p \right) +
	\nabla_{\cP_k} f\left( x \right)\right)^\top p_{\cP_k} + \hat
	\Psi_k\left( p_{\cP_k}\right).
\end{equation*}
From the boundedness of $H$, one can easily prove that
\eqref{eq:accept} is satisfied after a finite number of increases of
$\psi_i$, as we will show in Section~\ref{sec:analysis}.  In our
algorithm, SpaRSA runs until either a fixed number of iterations is
reached, or when some certain inner stopping condition for optimizing
\eqref{eq:quadratic} is satisfied.

For general $H$, the computational bottleneck of $\nabla \hat f$ would
take $O(N^2)$ operations to compute the $Hp^{(i)}$ term. However,
for our LBFGS choice of $H$, this cost is reduced to $O(mN + m^2)$
by utilizing the matrix structure, as shown in the following formula:
\begin{align}
\nabla \hat f\left(p\right) = \nabla f\left( x \right)
+ H p
= \nabla f(x) + \gamma p - U_t \left(M_t^{-1} \left(U_t^\top
p\right)\right).
\label{eq:u}
\end{align}
The computation of \eqref{eq:u} can be parallelized, by first
parallelizing computation of the inner product $U_t^\top p^{(i)}$ via
the formula
\begin{equation*}
	\sum_{k=1}^K \left(U_t\right)_{\cP_k,:}^\top p^{(i)}_{\cP_k}
\end{equation*}
with $O(m)$ communication.
(We implement the parallel inner products as described in
Section~\ref{subsec:products}.)
We let each machine compute a subvector of $u$ in \eqref{eq:udef}
according to \eqref{eq:psiblockseparable}.

From the block-separability of $\Psi$, the subproblem \eqref{eq:dk}
for computing $p^{(i)}$ can be decomposed into independent subproblems
partitioned along $\cP_1,\dotsc,\cP_K$. The $k$th machine therefore
locally computes $p^{(i)}_{\cP_k}$ without communicating the whole
vector. Then at each iteration of SpaRSA, partial inner products
between $(U_t)_{\cP_k,:}$ and $p^{(i)}_{\cP_k}$ can be computed
locally, and the results are assembled with an allreduce operation of
$O(m)$ communication cost.
This leads to a round of $O(m)$ communication cost per SpaRSA
iteration, with the computational cost reduced from $O(mN)$ to
$O(mN/K)$ per machine on average. Since both the $O(m)$ communication
cost and the $O(mN/K)$ computational cost are inexpensive when $m$ is
small, in comparison to the computation of $\nabla f$, one can afford
to conduct multiple iterations of SpaRSA at every main iteration. Note
that the total latency incurred over all allreduce operations as
discussed in \eqref{eq:commcost} can be capped by setting a maximum
iteration limit for SpaRSA.

The distributed implementation of SpaRSA for solving \eqref{eq:quadratic}
is summarized in Algorithm~\ref{alg:sparsa}.

\begin{algorithm}[tb]
	\DontPrintSemicolon
\caption{Distributed SpaRSA for solving \eqref{eq:quadratic} with
LBFGS quadratic approximation \eqref{eq:Hk} on machine $k$}
\label{alg:sparsa}
\begin{algorithmic}[1]
\STATE Given $\beta, \sigma_0 \in (0,1)$,
$M_t^{-1}$, $U_t$,  $\gamma_t$, and $\cP_k$;
\STATE Set $p^{(0)}_{\cP_k} \leftarrow 0$;
	\FOR{$i=0,1,2,\dotsc$}
		\IF{$i=0$}
			\STATE $\psi = \gamma_t$;
		\ELSE
			\STATE Compute $\psi$ in \eqref{eq:psi} through
			\Comment*[r]{$O(1)$ communication}
			\vspace{-10pt}
			\begin{equation*}
				\sum_{j=1}^K \left(p^{(i)}_{\cP_j} -
				p^{(i-1)}_{\cP_j}\right)^\top
				\left(\nabla_{\cP_j} \hat f \left(p^{(i)} \right) - \nabla_{\cP_j}
				\hat f \left(p^{(i-1)} \right)\right),\quad\text{ and }\quad
				\sum_{j=1}^K
				\left\|p^{(i)}_{\cP_j} - p^{(i-1)}_{\cP_j}\right\|^2;
			\end{equation*}
		\ENDIF
		\STATE Obtain \Comment*[r]{$O(m)$ communication}
		\vspace{-10pt}
		\begin{equation*}
			U_t^\top p^{(i)} = \sum_{j=1}^K \left(U_t\right)_{\cP_j,:}^\top
			p^{(i)}_{\cP_j};
		\end{equation*}
		\STATE Compute \[\nabla_{\cP_k} \hat
		f\left(p^{(i)}\right) = \nabla_{\cP_k} f \left(x\right) + \gamma
		p^{(i)}_{\cP_k} -
	\left(U_t\right)_{\cP_k,:}\left(M_t^{-1} \left(U_t^\top
p^{(i)}\right)\right)\]
by \eqref{eq:u};
	\STATE Solve \eqref{eq:dk} on coordinates indexed by $\cP_k$ to
				obtain $p_{\cP_k}$;
		\WHILE{TRUE}
			\IF{\eqref{eq:accept} holds \Comment*[r]{$O(1)$ communication}}
			\STATE $p^{(i+1)}_{\cP_k} \leftarrow p_{\cP_k}$; $\psi_i \leftarrow \psi$;
			\STATE Break;
			\ENDIF
			\STATE $\psi \leftarrow \beta^{-1} \psi$;
			\STATE Re-solve \eqref{eq:dk} with the new $\psi$ to
			obtain a new $p_{\cP_k}$;
		\ENDWHILE
		\STATE Break if some stopping condition is met;
	\ENDFOR
\end{algorithmic}
\end{algorithm}

\subsection{Sufficient Function Decrease}
After obtaining an update direction $p$ by approximately solving
\eqref{eq:quadratic}, we need to ensure sufficient objective decrease.
This is usually achieved by some line-search or trust-region procedure.
In this section, we describe two such approaches, one based on
backtracking line search for the step size, and one based on a
trust-region like approach that modifies $H$ repeatedly until an update
direction is accepted with unit step size.

For the line-search approach, we follow \citet{TseY09a} by using a
modified-Armijo-type backtracking line search to find a suitable step
size $\lambda$.  Given the current iterate $x$, the update direction
$p$, and parameters $\sigma_1, \theta \in (0,1)$, we set
\begin{equation}
\Delta \coloneqq \nabla f\left( x \right)^\top p + \Psi\left(x + p
\right) - \Psi\left(x \right)
\label{eq:Delta}
\end{equation}
and pick the step size as the largest of $\theta^0, \theta^1,\dotsc$
satisfying
\begin{equation}
F\left(x + \lambda p \right) \leq F\left( x \right) + \lambda
\sigma_1 \Delta.
\label{eq:armijo}
\end{equation}
The computation of $\Delta$ is negligible
as all the terms are involved in $Q(p;x)$, and $Q(p;x)$ is evaluated in
the line search procedure of SpaRSA.
For the function value evaluation, the objective values of both
\eqref{eq:primal} and \eqref{eq:dual} can be evaluated efficiently if
we precompute $Xp$ or $X^\top p$ in advance and conduct all reevaluations
through this vector but not repeated matrix-vector products.  Details
are discussed in Section \ref{sec:primaldual}.
Note that because $H_t$ defined in \eqref{eq:Hk} attempts to
approximate the real Hessian, empirically the unit step $\lambda=1$
frequently satisfies \eqref{eq:armijo}, so we use the value $1$ as the
initial guess.

For the trust-region-like procedure, we start from the original $H$,
and use the same $\sigma_1, \theta \in (0,1)$ as above.
Whenever the sufficient decrease condition
\begin{equation}
	F\left( x + p \right) - F\left( x \right) \leq \sigma_1 Q_H(p;x)
	\label{eq:decrease}
\end{equation}
is not satisfied, we scale up $H$ by $H \leftarrow H/\theta$, and
resolve \eqref{eq:quadratic}, either from $0$ or from the previously
obtained solution $p$ if it gives an objective better than $0$.
We note that when $\Psi$ is not present, both the backtracking
approach and the trust-region one generate the same iterates. But when
$\Psi$ is incorporated, the two approaches may generate different updates.
Similar to the line-search approach, the evaluation of $Q_H(p;x)$
comes for free from the SpaRSA procedure, and usually the original $H$
\eqref{eq:Hk} generates update steps satisfying \eqref{eq:decrease}.
Therefore, solving \eqref{eq:quadratic} multiple times per main
iteration is barely encountered in practice.

The trust-region procedure may be more expensive than line search because
solving the subproblem again is more expensive than trying a
different step size, although both cases are empirically rare.
But on the other hand, when there are additional properties of the
regularizer such as sparsity promotion, a potential benefit of
the trust-region approach is that it might be able to identify the
sparsity pattern earlier because unit step size is always used.

Our distributed algorithm for \eqref{eq:f} is summarized in
Algorithm~\ref{alg:proximalbfgs}. We refer to the line search and
trust-region variants of the algorithm as DPLBFGS-LS and DPLBFGS-TR
respectively, and we will refer to them collectively as simply DPLBFGS.
\begin{algorithm}[tb]
	\DontPrintSemicolon
\caption{DPLBFGS: A distributed proximal variable-metric LBFGS method for \eqref{eq:f}}
\label{alg:proximalbfgs}
\begin{algorithmic}[1]
\STATE Given $\theta, \sigma_1 \in (0,1)$, $\delta > 0$, an initial
point $x=x^0$, a partition $\{\cP_k\}_{k=1}^K$ satisfying \eqref{eq:psiblockseparable};
\FOR{Machines $k=1,\dotsc,K$ in parallel}
\STATE Obtain $F(x)$; \Comment*[r]{$O(1)$ communication}
	\FOR{$t=0,1,2,\dotsc$}
		\STATE Compute $\nabla f(x)$; \Comment*[r]{$O(d)$
			communication}
		\STATE Initialize $H$; 
		\IF{$t \neq 0$ and \eqref{eq:safeguard} holds for $(\bs_{t-1},
		\by_{t-1})$ \Comment*[r]{$O(1)$ communication}}
			\STATE Update $U_{\cP_k,:}$, $M$, and $\gamma$ by
				\eqref{eq:M}-\eqref{eq:updates};
				\Comment*[r]{$O(m)$ communication}
			\STATE Compute $M^{-1}$;
			\STATE Implicitly form a new $H$ from \eqref{eq:Hk};
		\ENDIF
		\IF{$U$ is empty}
			\STATE Solve \eqref{eq:quadratic} using some existing
			distributed algorithm to obtain $p_{\cP_k}$;
		\ELSE
			\STATE Solve \eqref{eq:quadratic} using
				Algorithm~\ref{alg:sparsa} in a distributed manner
				to obtain $p_{\cP_k}$;
		\ENDIF
		\IF{Line search}
			\STATE Compute $\Delta$ defined in \eqref{eq:Delta};
			\FOR{$i=0,1,\dotsc$}
				\STATE $\lambda = \theta^i$;
				\STATE Compute $F(x + \lambda p)$; \Comment*[r]{$O(1)$
					communication}
				\IF{$F(x + \lambda p) \leq F(x) + \sigma_1 \lambda \Delta$}
					\STATE Break;
				\ENDIF
			\ENDFOR
		\ELSIF{Trust region}
			\STATE $\lambda = 1$;
			\STATE Compute $Q_H(p;x)$;
			\WHILE{$F(x + p) - F(x) > \sigma_1 Q_H(p;x)$
			\Comment*[r]{$O(1)$ communication}}
				\STATE $H \leftarrow H / \theta$;
				\STATE Re-solve \eqref{eq:quadratic} to obtain update
					$p_{\cP_k}$;
				\STATE Compute $Q_H(p;x)$;
			\ENDWHILE
		\ENDIF
		\STATE $x_{\cP_k} \leftarrow x_{\cP_k} +\lambda p_{\cP_k}$, $F(x)
			\leftarrow F(x + \lambda p)$;
		\STATE $x^{t+1} \coloneqq x$;
		\STATE $(\bs_t)_{\cP_k} \leftarrow x^{t+1}_{\cP_k} -
			x^t_{\cP_k}$, $(\by_t)_{\cP_k} \leftarrow \nabla_{\cP_k}
			f(x^{t+1} - \nabla_{\cP_k} f (x^t)$;
	\ENDFOR
\ENDFOR
\end{algorithmic}
\end{algorithm}

\subsection{Cost Analysis}
\label{subsec:cost}
We now describe the computational and communication cost of our
algorithms. The computational cost for each machine depends on which
$X_k$ is stored locally and the size of $|\cP_k|$, and for simplicity
we report the computational cost \emph{averaged over all machines}.
The communication costs do not depend on $X_k$.

For the distributed version of Algorithm
\ref{alg:sparsa},
each iteration costs
\begin{equation}
	O\left(\frac{N}{K} + \frac{mN}{K} +  m^2\right) =
	O\left(\frac{mN}{K} + m^2\right)
	\label{eq:costsparsa}
\end{equation}
in computation,
where the $N/K$ term is for the vector additions in \eqref{eq:u},
and
\begin{equation*}
	O\left(m + \text{number of times \eqref{eq:accept} is
evaluated}\right)
\end{equation*}
in communication.
In the next section, we will show that \eqref{eq:accept} is accepted
within a fixed number of times and thus the overall communication cost
is $O(m)$.

For DPLBFGS, we will give details in Section
\ref{sec:primaldual} that for both \eqref{eq:primal} and
\eqref{eq:dual}, each gradient evaluation for $f$ takes
$O(\#\text{nnz} / K)$ per machine computation in average and $O(d)$ in communication, where \#nnz is the number of
nonzero elements in the data matrix $X$.
As shown in the next section, in one main iteration, the number
of function evaluations in the line search is bounded, and its cost is
negligible if we are using the same $p$ but just different step sizes;
see Section \ref{sec:primaldual}.
For the trust region approach,
the number of times for modifying $H$ and resolving
\eqref{eq:quadratic} is also bounded,
and thus the asymptotical cost is not altered.
In summary, the computational cost per main iteration is therefore
\begin{equation}
	O\left(\frac{\#\text{nnz}}{K} + \frac{mN}{K} +
	m^3 + \frac{N}{K}\right) = O\left( \frac{\#\text{nnz}}{K} +
	\frac{mN}{K} + m^3 \right),
	\label{eq:costmain}
\end{equation}
and the communication cost is
\begin{equation*}
	O\left( 1 + d \right) = O\left( d \right),
\end{equation*}
where the $O(1)$ part is for function value evaluation and checking
the safeguard \eqref{eq:safeguard}.
We note that the costs of Algorithm \ref{alg:sparsa} are
dominated by those of DPLBFGS if a fixed number of SpaRSA
iterations is conducted every main iteration.

\section{Convergence Rate and Communication Complexity Analysis}
\label{sec:analysis}
The use of an iterative solver for the subproblem \eqref{eq:quadratic}
generally results in an inexact solution.  We first show that running
SpaRSA for any fixed number of iterations guarantees a step $p$
whose accuracy is sufficient to prove overall convergence.

\begin{lemma}
\label{lemma:sparsa}
Consider optimizing \eqref{eq:f} by DPLBFGS.
By using $H_t$ as defined in \eqref{eq:Hk} with the safeguard mechanism
\eqref{eq:safeguard} in \eqref{eq:quadratic}, we have the following.
\begin{compactenum}
\item
We have $L^2/\delta \geq \gamma_t \geq \delta$ for all
$t > 0$, where $L$ is the Lipschitz constant for $\nabla f$.
Moreover, there exist constants $c_1 \geq c_2 > 0$ such that $c_1 I
\succeq H_t \succeq c_2 I$  for all $t>0$.
\item
At every SpaRSA iteration, the initial estimate of $\psi_i$
is bounded within the range of
$$\left[\min\left\{c_2, \delta\right\}, \max\left\{c_1,
\frac{L^2}{\delta}\right\}\right],$$ and the final
accepted value $\psi_i$ is upper-bounded.
\item
SpaRSA is globally Q-linear convergent in solving
\eqref{eq:quadratic}.
Therefore, there exists $\eta \in [0,1)$ such that if we run
	at least $S$ iterations of SpaRSA for all main iterations for any
	$S>0$, the approximate solution $p$ satisfies
\begin{equation}
-\eta^S Q^* = \eta^S\left(Q\left(0\right) - Q^*\right) \geq
Q\left(p\right) - Q^*,
\label{eq:approx}
\end{equation}
where $Q^*$ is the optimal objective of
\eqref{eq:quadratic}.
\end{compactenum}
\end{lemma}

Lemma \ref{lemma:sparsa} establishes how the number of iterations of
SpaRSA affects the inexactness of the subproblem solution.
Given this measure,
we can leverage the results developed in \citet{LeeW18a,PenZZ18a} to
obtain iteration complexity guarantees for our algorithm. Since in our
algorithm, communication complexity scales linearly with iteration
complexity, this guarantee provides a bound on the amount of
communication.
In particular, our method communicates
$O(d+mS)$ bytes per iteration (where $S$ is the number of SpaRSA
iterations used, as in Lemma~\ref{lemma:sparsa}) and the second term
can usually be ignored for small $m$.

We show next that the step size generated by our line search procedure
in DPLBFGS-LS is lower bounded by a positive value.
\begin{lemma}
\label{lemma:delta}
Consider \eqref{eq:f} such that $f$ is $L$-Lipschitz differentiable
and $\Psi$ is convex.
If SpaRSA is run at least $S$ iterations in solving
\eqref{eq:quadratic},
the corresponding $\Delta$ defined in \eqref{eq:Delta} satisfies
\begin{equation}
	\Delta \leq -\frac{c_2 \left\|p\right\|^2}{
	1 + \eta^{\frac{S}{2}}},
\label{eq:deltabound}
\end{equation}
where $\eta$ and $c_2$ are the same as that defined in Lemma \ref{lemma:sparsa}.
Moreover, the backtracking subroutine in DPLBFGS-LS terminates in
finite number of steps and produces a step size
\begin{equation}
	\lambda \geq \min\left\{1, \frac{2\theta\left(1 -
\sigma_1\right) c_2}{L\left(1 + \eta^{{S}/{2}}\right)}\right\}
\label{eq:linesearchbound2}
\end{equation}
satisfying \eqref{eq:armijo}.
\end{lemma}
We also show that for the trust-region technique, at one main
iteration, the number of times we solve the subproblem
\eqref{eq:quadratic} until a step is accepted is upper-bounded by a
constant.
\begin{lemma}
\label{lemma:Hbound}
For DPLBFGS-TR, suppose each time when we solve \eqref{eq:quadratic} we
have guarantee that the objective value is no worse than $Q(0)$. Then
when \eqref{eq:decrease} is satisfied, we have that
\begin{equation}
\|H_t\| \leq c_1 \max\left\{1, \frac{L
}{c_2 \theta}\right\}.
\label{eq:m1}
\end{equation}
Moreover, at each main iteration, the number of times we solve
\eqref{eq:quadratic} with different $H$ is upper-bounded by
\begin{equation*}
	\max\left\{ 1, \left\lceil\log_{\theta} \frac{c_2}{L}\right\rceil \right\}
\end{equation*}
\end{lemma}
Note that the bound in Lemma \ref{lemma:Hbound} is independent to the
number of SpaRSA iterations used.
It is possible that one can incorporate the subproblem suboptimality
to derive tighter but more complicated bounds,
but for simplicity we use the current form of Lemma
\ref{lemma:Hbound}.

The results in Lemmas \ref{lemma:delta}-\ref{lemma:Hbound} are just
worst-case guarantees; in practice we often observe that the line
search procedure terminates with $\lambda=1$ for our original choice
of $H$, as we see in our experiments.  This also indicates that in
most of the cases, \eqref{eq:decrease} is satisfied with the original
LBFGS Hessian approximation without scaling $H$.

We now lay out the main theoretical results in
Theorems~\ref{thm:Fconvex} to \ref{thm:Fnonconvex}, which describe the
iteration and communication complexity under different conditions on
the function $F$. In all these results, we assume the following
setting:
\begin{quote}
We apply DPLBFGS to solve the main problem \eqref{eq:f}, running
Algorithm~\ref{alg:sparsa} for $S$ iterations in each main iteration.
Let $x^t$, $\lambda_t$, and $H_t$ be respectively the $x$ vector, the
step size, and the final accepted quadratic approximation matrix at the
$t$th iteration of DPLBFGS for all $t\geq 0$. Let $M$ be the supremum
of $\|H_t\|$ for all $t$ (which is either $c_1$ or $c_1 L / (c_2
\theta)$ according to Lemmas~\ref{lemma:sparsa} and
\ref{lemma:Hbound}), and $\bar\lambda$ be the infimum of the step
sizes over iterations (either $1$ or the bound from Lemma
\ref{lemma:delta}). Let $F^*$ be the optimal objective value of
\eqref{eq:f}, $\Omega$ the solution set, and $P_{\Omega}$ the (convex)
projection onto $\Omega$.
\end{quote}

\begin{theorem} \label{thm:Fconvex}
If $F$ is convex, given an initial point $x^0$, assume
\begin{equation}
R_0 \coloneqq \sup_{x: F\left( x \right) \leq F\left( x^0
	\right)}\quad\left\|x - P_\Omega(x)\right\|
	\label{eq:R0}
\end{equation}
is finite, we obtain the following expressions for rate of convergence
of the objective value.
\begin{compactenum}
\item
	When $$F(x^t) - F^* \geq \left( x^t - P_{\Omega}\left( x^t
		\right) \right)^\top H_t \left( x^t - P_{\Omega}\left(
		x^t\right)\right),$$ the objective converges linearly to the
		optimum:
\begin{equation*}
\frac{F(x^{t+1}) - F^*}{F(x_t)^- F^*} \leq 1 - \frac{\left(
	1 - \eta^S \right)\sigma_1 \lambda_t}{2}.
\end{equation*}
\item For any $t \geq t_0$,
	where
	\begin{equation*}
		t_0 \coloneqq \arg \min\{t\mid M R_0^2 > F\left( x^t
		\right) - F^*\},
	\end{equation*}
	we have
\begin{align*}
	F\left(x^t\right) - F^* &\leq \frac{2 M R_0^2}{\sigma_1 (1 -
		\eta^S)\sum_{i=t_0}^{t-1} \lambda_t + 2}.
\end{align*}
Moreover,
\begin{equation*}
	t_0 \leq \max\left\{ 0, 1 + \frac{2}{\sigma_1 (1 - \eta^s)
		\bar{\lambda}} \log \frac{f\left( x^0 \right) - f^*}{M R_0^2}
	\right\}.
\end{equation*}
\end{compactenum}
Therefore, for any $\epsilon > 0$, the number of rounds of $O(d)$
communication required to obtain an $x^t$ such that $F(x^t) - F^* \leq
\epsilon$ is at most
\begin{equation*}
	\begin{cases}
	O\left(\max\left\{ 0, 1 + \frac{2}{\sigma_1 (1 - \eta^s)
		\bar{\lambda}} \log \frac{F\left( x^0 \right) - F^*}{M
		R_0^2}\right\}
		+ \frac{2 M R_0^2}{\sigma_1 \bar{\lambda} \left( 1 - \eta^S \right)\epsilon}
		\right)&\text{ if } \epsilon < M R_0^2,\\
		O\left(\max\left\{ 0, 1 + \frac{2}{\sigma_1 \left( 1 - \eta^S \right)
			\bar{\lambda}} \log \frac{F\left( x^0 \right) -
			F^*}{\epsilon} \right\}\right)
			&\text{ else}.
	\end{cases}
\end{equation*}
\end{theorem}

\begin{theorem}
When $F$ is convex and the quadratic growth condition
	\begin{equation}
		F\left( x \right) - F^* \geq \frac{\mu}{2} \left\| x -
		P_{\Omega}\left( x \right) \right\|^2, \quad \forall x \in \R^N
		\label{eq:qg}
	\end{equation}
	holds for some $\mu > 0$, we get a global Q-linear convergence
	rate:
\begin{equation}
\frac{F\left(x^{t+1}\right) - F^*}{F\left( x^t \right) - F^*}
\leq 1 - \lambda_t \sigma_1 \left(1 - \eta^S\right)\cdot
\begin{cases}
	\frac{\mu}{4\|H_t\|}, &\text{ if } \mu \leq 2 \|H_t\|,\\
	1 - \frac{\|H_t\|}{\mu}, &\text{ else.}
\end{cases}
\label{eq:qlinear}
\end{equation}
Therefore, the rounds of $O(d)$ communication needed for getting an
$\epsilon$-accurate objective is
\begin{equation*}
\begin{cases}
O\left(\max\left\{ 0, 1 + \frac{2}{\sigma_1 (1 - \eta^s)
	\bar{\lambda}} \log \frac{F\left( x^0 \right) - F^*}{M
	R_0^2}\right\}
	+ \frac{4M}{\mu \bar{\lambda}\sigma_1 \left( 1 - \eta^S \right)}
		\log\frac{M R_0^2}{\epsilon}
	\right)&\text{ if } \epsilon < M R_0^2, \mu \leq 2 M,\\
O\left(\max\left\{ 0, 1 + \frac{2}{\sigma_1 (1 - \eta^s)
	\bar{\lambda}} \log \frac{F\left( x^0 \right) - F^*}{M
	R_0^2}\right\}
	+ \frac{\mu}{(\mu - M) \bar{\lambda}\sigma_1 \left( 1 - \eta^S \right)}
		\log\frac{M R_0^2}{\epsilon}
	\right)&\text{ if } \epsilon < M R_0^2, \mu > 2 M,\\
	O\left( 0, 1 + \frac{2}{\sigma_1 \left( 1 - \eta^S \right)
		\bar{\lambda}} \log \frac{F\left( x^0 \right) - F^*}{\epsilon} \right)
		&\text{ if } \epsilon \geq MR_0^2.
\end{cases}
\end{equation*}
\end{theorem}

\begin{theorem}
Suppose that the following relaxation of strong convexity holds:
	There exists $\mu > 0$ such that for any $x\in \R^N$ and any $a
	\in [0,1]$, we have
\begin{equation}
F\left(a x + \left(1 - a\right)
P_{\Omega}\left(x\right)\right)
\leq a F\left(x\right) +
\left(1 - a\right) F^* - \frac{\mu a\left( 1 - a \right)}{2}
\left\|x - P_{\Omega}\left(x\right)\right\|^2.
\label{eq:strong}
\end{equation}
Then DPLBFGS converges globally at a
Q-linear rate faster than \eqref{eq:qlinear}. More specifically,
\begin{align*}
\frac{F\left(x^{t+1}\right) - F^*}{F\left( x^t \right) - F^*}
\leq
1 - \frac{\lambda_t \sigma_1 \left(1 -\eta^S \right)\mu}{\mu +
\|H_t\|}.
\end{align*}
Therefore, to get an approximate solution of \eqref{eq:f} that
is $\epsilon$-accurate in the sense of objective value, we need to
perform at most
\begin{equation*}
	\begin{cases}
	O\left(\max\left\{ 0, 1 + \frac{2}{\sigma_1 (1 - \eta^s)
		\bar{\lambda}} \log \frac{F\left( x^0 \right) - F^*}{M
		R_0^2}\right\} +
	\frac{\mu + M}{\mu \sigma_1 \bar{\lambda} \left( 1 -
\eta^S \right)} \log\frac{M R_0^2}{\epsilon} \right)
		&\text{ if } \epsilon < M R_0^2,\\
		O\left( 0, 1 + \frac{2}{\sigma_1 \left( 1 - \eta^S \right)
			\bar{\lambda}} \log \frac{F\left( x^0 \right) - F^*}{\epsilon} \right)
			&\text{ else}.
	\end{cases}
\end{equation*}
rounds of $O(d)$ communication.
\end{theorem}

\begin{theorem} \label{thm:Fnonconvex}
 If $F$ is non-convex, the norm of
\begin{equation*}
G_t \coloneqq \arg\min_p \quad \nabla f\left(x^t\right)^\top p +
\frac{\|p\|^2}{2}
+ \Psi\left(x + p\right)
\end{equation*}
converges to zero at a rate of $O(1 / \sqrt{t})$ in the following sense:
\begin{align*}
\min_{0 \leq i \leq t }\left\|G_i\right\|^2 \leq \frac{F\left( x^0
	\right) - F^*}{\sigma_1 \left( t+1 \right)} \frac{M^2\left(1 +
	\frac{1}{c_2} + \sqrt{1 - \frac{2}{M} +
	\frac{1}{c_2^2}}\right)^2}{2 c_2(1 - \eta^S) \min_{0\leq i \leq t}
\lambda_i }.
\end{align*}
Moreover, if there are limit points in the sequence $\{x^0,
x^1,\dotsc\}$, then all limit points are stationary.
\end{theorem}
Note that it is known that the norm of $G_t$ is zero if and only if
$x^t$ is a stationary point, so this measure serves as an indicator
for the first-order optimality condition.
The class of quadratic growth \eqref{eq:qg} includes many
non-strongly-convex ERM problems.
Especially, it contains problems of the form
\begin{equation}
	\min_{x\in \mathcal X}\quad g\left( Ax \right) + b^\top x,
	\label{eq:qgform}
\end{equation}
where $g$ is strongly convex, $A$ is a matrix, $b$ is a vector, and
$\mathcal{X}$ is a polyhedron.
Commonly seen non-strongly-convex ERM problems including
$\ell_1$-regularized logistic regression, LASSO, and the dual problem
of support vector machines all fall in the form \eqref{eq:qgform} and
therefore our algorithm enjoys global linear convergence on them.


\section{Solving the Primal and the Dual Problem}
\label{sec:primaldual}
Now we discuss details on how to apply DPLBFGS described in the
previous section to the specific problems
\eqref{eq:primal} and \eqref{eq:dual} respectively.
We discuss how to obtain the gradient of the smooth part $f$ and how to
conduct line search efficiently under distributed data storage.
For the dual problem, we additionally describe how to recover a
primal solution from our dual iterates.

\subsection{Primal Problem}
\label{subsec:primal}
Recall that the primal problem is \eqref{eq:primal}
$
\min_{\bw \in \R^d} \xi\left( X^\top \bw \right) + g\left( \bw \right),
$
and is obtained from the general form \eqref{eq:f} by having $N =d$,
$x = \bw$, $f(\cdot) = \xi(X^\top \cdot)$, and $\Psi(\cdot) =
g(\cdot)$. The gradient of $\xi$ with respect to $\bw$ is 
\begin{equation*}
X \nabla \xi(X^\top \bw) = \sum_{k=1}^K \left(X_k
\nabla \xi_k (X_k^\top \bw)\right).
\end{equation*}
We see that, except for the sum over $k$, the computation can be
conducted locally provided $\bw$ is available to all machines. Our
algorithm maintains $X_k^\top \bw$ on the $k$th machine throughout, and
the most costly steps are the matrix-vector multiplications between
$X_k$ and $\nabla \xi_k(X_k^\top \bw)$.
Clearly, computing $X_k^\top \bw$ and $X_k \nabla
\xi_k(X_k^\top \bw)$ both cost $O(\#\text{nnz} / K)$ in average among
the $K$ machines.  The local $d$-dimensional partial gradients are
then aggregated through an allreduce operation using a round of $O(d)$
communication.

To initialize the approximate Hessian matrix $H$ at $t=0$, we set $H_0
\coloneqq a_0 I$ for some positive scalar $a_0$.
In particular,
we use
\begin{equation}
a_0 \coloneqq \frac{\left|\nabla f(\bw_0)^\top \nabla^2 f(\bw_0) \nabla
	f(\bw_0)\right|}{\left\|\nabla f(\bw_0)\right\|^2},
\label{eq:a0}
\end{equation}
where $\nabla^2 f(\bw_0)$ denotes the generalized Hessian when $f$ is
not twice-differentiable.

For the function value evaluation part of line search, each machine
will compute $\xi_k(X_k^\top\bw + \lambda X_k^\top) +
g_k(\bw_{\cPg_k} + \lambda p_{\cPg_k})$
(the left-hand side of \eqref{eq:armijo}) and send this scalar over the
network. Once we have precomputed $ X_k^\top \bw$ and $X_k^\top p$, we 
can quickly obtain $X_k^\top(\bw + \lambda p)$ for any value of $\lambda$
without having to performing matrix-vector multiplications.
Aside from the communication needed to compute the summation of the
$f_k$ terms in the evaluation of $f$, the only other communication
needed is to share the update direction $p$ from subvectors
$p_{\cPg_k}$.
Thus, two rounds of $O(d)$ communication are incurred per main
iteration.

\subsection{Dual Problem}
\label{subsec:dual}
Now consider applying DPLBFGS to the dual problem \eqref{eq:dual}.
To fit it into the general form \eqref{eq:f},
we have $N = n$, $x = \AL$, $f(\cdot) = g^*(X\cdot)$, and $\Psi(\cdot)
= -\xi^*(-\cdot)$.
In this case, we need a way to efficiently obtain the vector $$\bz
\coloneqq X\AL$$
on each machine in order to compute $g^*\left( X\AL \right)$ and
the gradient $X^\top \nabla g^*(X\AL)$. 

Since each machine has access to some columns of $X$, it is natural to
split $\AL$ according to the same partition. Namely, we set $\cP_k$ as
described in \eqref{eq:psiblockseparable} to $\cPX_k$. Every
machine can then individually compute $X_k \AL_k$, and after one round
of $O(d)$ communication, each machine has a copy of $\bz = X\AL =
\sum_{k=1}^K X_k \AL_k$. After using $\bz$ to compute $\nabla_{\bz}
g^*(\bz)$, we can compute the gradient $\nabla_{\cPX_k} g^*(X\AL) =
X_k^\top \nabla g^*(X\AL)$ at a computation cost of $O(\#\text{nnz}/K)$
in average among the $K$ machines, matching the cost of computing $X_k
\AL_k$ earlier.


To construct the approximation matrix $H_0$ for the first main
iteration, we make use of the fact that the (generalized) Hessian of
$g^*(X\AL)$ is
\begin{equation}
	X^\top \nabla^2 g^*(\bz) X.
	\label{eq:Hess}
\end{equation}
Each machine has access to one $X_k$, so we can construct the
block-diagonal proportion of this Hessian locally for the part
corresponding to $\cPX_k$. Therefore, the block-diagonal part of the
Hessian is a natural choice for $H_0$. Under this
choice of $H_0$, the subproblem \eqref{eq:quadratic} is
decomposable along the $\cPX_1,\dotsc,\cPX_K$ partition and one can
apply algorithms other than SpaRSA to solve this. For example, we can
apply CD solvers on the independent local subproblems, as done by
\cite{LeeC17a, Yan13a, ZheXXZ17a}.
As it is observed in these works that the block-diagonal approaches
tend to converge fast at the early iterations, we use it for
initializing our algorithm.
In particular, we start with the block-diagonal approach, until $U_t$
has $2m$ columns, and then we switch to the LBFGS approach.
This turns out to be much more efficient in practice than starting
with the LBFGS matrix.

For the line search process, we can precompute the matrix-vector
product $Xp$ with the same $O(d)$ communication and $O(\#\text{nnz}/K)$
per machine average computational cost as computing $X\AL$. 
With $X\AL$ and $Xp$, we can now evaluate $X\AL
+ \lambda Xp$ quickly for different $\lambda$, instead of having to
perform a matrix-vector multiplication of the form $X(\AL + \lambda p)$
for every $\lambda$.
For most common choices of $g$, given $\bz$, the computational cost of
evaluating $g^*(\bz)$ is $O(d)$.
Thus, the cost of this efficient implementation per backtracking
iteration is reduced to $O(d)$, with an overhead of
$O(\#\text{nnz}/K)$ per machine average per main iteration, while the
naive implementation takes $O(\#\text{nnz}/K)$ per backtracking
iteration.
After the sufficient decrease condition holds, we locally update
$\AL_k$ and $X\AL$ using $p_{\cPX_k}$ and $Xp$.
For the trust region approach, the two implementations take the same
cost.


\subsubsection{Recovering a Primal Solution}


In general, the algorithm only gives us an approximate solution to
the dual problem \eqref{eq:dual}, which means the formula 
\begin{equation}
  \bw\left(\AL\right) \coloneqq \nabla g^*\left( X \AL \right).
  \label{eq:walpha}
\end{equation}
used to obtain a primal optimal point from a dual optimal point
(equation \eqref{eq:wopt}, derived from KKT conditions) is no longer
guaranteed to even return a feasible point without further assumptions.
Nonetheless, this is a common approach and under certain conditions
(the ones we used in Assumption \ref{assum:dual}), one can provide
guarantees on the resulting point.

It can be shown from existing works \citep{Bac15a, ShaZ12a} that when
$\AL$ is not an optimum for \eqref{eq:dual}, for \eqref{eq:walpha},
certain levels of primal suboptimality can be achieved,
which depend on whether $\xi$ is Lipschitz-continuously differentiable
or Lipschitz continuous.
This is the reason why we need the corresponding assumptions in
Assumption \ref{assum:dual}.
A summary of those results is available in \cite{LeeC17a}. We restate
their results here for completeness but omit the proof.

\begin{theorem}[{\citet[Theorem 3]{LeeC17a}}]
\label{thm:dualitygap}
Given any $\epsilon > 0$, and
any dual iterate $\AL \in \R^n$ satisfying
\begin{equation*}
	D(\AL) - \min_{\bar\AL \in \R^n} \quad D(\bar\AL) \leq \epsilon.
\end{equation*}
If Assumption \ref{assum:dual} holds,
then the following results hold.
\begin{compactenum}
\item  If the part in Assumption \ref{assum:dual} that $\xi^*$ is
	$\sigma$-strongly convex holds, then $\bw(\AL)$ satisfies
	\begin{equation*}
		P\left( \bw\left( \AL \right) \right) - \min_{\bw\in
		\R^d}\quad P\left( \bw \right) \leq \epsilon \left( 1 +
		\frac{L}{\sigma} \right).
	\end{equation*}
\item  If the part in Assumption \ref{assum:dual} that $\xi$ is
	$\rho$-Lipschitz continuous holds, then $\bw(\AL)$ satisfies
	\begin{equation*}
		P\left( \bw\left( \AL \right) \right) - \min_{\bw\in
		\R^d}\quad P\left( \bw \right) \leq \max\left\{
			2 \epsilon, \sqrt{8 \epsilon \rho^2 L} \right\}.
	\end{equation*}
\end{compactenum}
\end{theorem}

One more issue to note from recovering the primal solution through
\eqref{eq:walpha} is that our algorithm only guarantees monotone
decrease of the dual objective but not the primal objective.
To ensure the best primal approximate solution,
one can follow \cite{LeeC17a} to maintain the primal iterate that gives
the best objective for \eqref{eq:primal} up to the current iteration
as the output solution. The theorems above still apply to this iterate
and we are guaranteed to have better primal performance.

\section{Related Works}
\label{sec:related}
The framework of using the quadratic approximation subproblem
\eqref{eq:quadratic} to generate update directions for optimizing
\eqref{eq:f} has been discussed in existing works with different
choices of $H$, but always in the single-core setting.
\citet{LeeSS14a} focused on using $H = \nabla^2 f$, and proved
local convergence results under certain additional assumptions.
In their experiment, they used AG to solve \eqref{eq:quadratic}.
However, in distributed environments, for \eqref{eq:primal} or
\eqref{eq:dual},
using $\nabla^2 f$ as $H$ needs an $O(d)$ communication per AG
iteration in solving \eqref{eq:quadratic}, because computation of the
term $\nabla^2 f(x) p$ involves either $X D X^\top p$ or $X^\top D
X p$ for some diagonal matrix $D$, which
requires one {\em allreduce} operation to calculate a weighted sum of
the columns of $X$.

\citet{SchT16a} and \citet{GhaS16a} showed global convergence rate
results for a method based on \eqref{eq:quadratic} with bounded $H$,
and suggested using randomized coordinate descent to solve
\eqref{eq:quadratic}.  In the experiments of these two works, they
used the same choice of $H$ as we do in this paper, with CD as the
solver for \eqref{eq:quadratic}, which is well suited to their
single-machine setting.  Aside from our extension to the distributed
setting and the use of SpaRSA, the third major difference between their
algorithm and ours is how sufficient objective decrease is guaranteed.
When the obtained solution with a unit step size does
not result in sufficient objective value decrease, they add a multiple
of the identity matrix to $H$ and solve \eqref{eq:quadratic} again
starting from $p^{(0)} = 0$.
This is different from how we modify $H$ and in some worst cases, the
behavior of their algorithm can be closer to a first-order method if
the identity part dominates, and more trials of different $H$ might be
needed.
The cost of repeatedly solving \eqref{eq:quadratic} from scratch can
be high, which results in an algorithm with higher overall complexity.
This potential inefficiency is exacerbated further by the inefficiency
of coordinate descent in the distributed setting.

Our method can be considered as a special case of the algorithmic
framework in \citet{LeeW18a, BonLPP16a}, which both focus on
analyzing the theoretical guarantees under various conditions for
general $H$. In the experiments of \citet{BonLPP16a}, $H$ is obtained
from the diagonal entries of $\nabla^2 f$, making the subproblem
\eqref{eq:quadratic} easy to solve, but this simplification does not
take full advantage of curvature information.
Although most our theoretical convergence analysis follows directly
from \citet{LeeW18a} and its extension \cite{PenZZ18a}, these works do
not provide details of experimental results or implementation, and
their analyses focus on general $H$ rather than the LBFGS choice we
use here.

For the dual problem \eqref{eq:dual}, there are existing distributed
algorithms under the instance-wise storage scheme (for example,
\cite{Yan13a, LeeC17a, ZheXXZ17a, DunLGBHJ18a} and the references
therein).  As we discussed in Section \ref{subsec:dual}, it is easy to
recover the block-diagonal part of the Hessian \eqref{eq:Hess} under
this storage scheme. Therefore, these works focus on using the
block-diagonal part of the Hessian and use \eqref{eq:quadratic} to
generate update directions. In this case, only blockwise curvature
information is obtained, so the update direction can be poor if the
data is distributed nonuniformly. In the extreme case in which each
machine contains only one column of $X$, only the diagonal entries of
the Hessian can be obtained, so the method reduces to a scaled version
of proximal gradient. Indeed, we often observe in practice that these
methods tend to converge quickly in the beginning, but after a while
the progress appears to stagnate even for small $K$.

\citet{ZheXXZ17a} give a primal-dual framework with acceleration
that utilizes a distributed solver for \eqref{eq:dual} to optimize
\eqref{eq:primal}.
Their algorithm is essentially the same as applying the Catalyst
framework \citep{LinMH15a} on a strongly-convex primal problem to form
an algorithm with an inner and an outer loop.
In particular, their approach consists of the following steps
per round to optimize a strongly-convex primal problem with the
additional requirement that $g$ being Lipschitz-continuously
differentiable.
\begin{compactenum}
\item Add a quadratic term centered at a given point $y$ to form a
subproblem with better condition.
\item Approximately optimize the new problem by using a distributed dual problem solver,
and
\item find the next $y$ through extrapolation techniques similar to that of
accelerated gradient \citep{Nes13a, BecT09a}.
\end{compactenum}
A more detailed description of the Catalyst framework (without
requiring both terms to be differentiable) is given in Appendix
\ref{app:catalyst}.
We consider one round of the above process as one outer iteration of
their algorithm, and the inner loop refers to the optimization process
in the second step.
The outer loop of their algorithm is conducted on the primal problem
\eqref{eq:primal} and a distributed dual solver is simply considered
as a subproblem solver using results similar to Theorem
\ref{thm:dualitygap}.
Therefore this approach is more a primal problem solver than a dual
one, and it should be compared with other distributed primal solvers
for smooth optimization but not with the dual algorithms.
However, the Catalyst framework can be applied directly on the dual
problem directly as well, and this type of acceleration can to some
extent deal with the problem of stagnant convergence appeared in the
block-diagonal approaches for the dual problem.
Unfortunately, those parameters used in acceleration are not just global
in the sense that the coordinate blocks are considered all together, but
also global bounds for all possible $\bw \in \R^d$ or $\AL \in \R^n$.
This means that the curvature information around the current iterate
is not considered, so the improved convergence can still be slow.
By using the Hessian or its approximation as in our method, we can get
much better empirical convergence.

A column-wise split of $X$ in the dual problem \eqref{eq:dual}
corresponds to a primal problem \eqref{eq:primal} where $X$ is split
row-wise.  Therefore, existing distributed algorithms for the dual ERM
problem \eqref{eq:dual} can be directly used to solve
\eqref{eq:primal} in a distributed environment where $X$ is
partitioned feature-wise (i.e. along rows instead of columns).
However, there are two potential disadvantages of this approach.
First, new data points can easily be assigned to one of the machines
in our approach, whereas in the feature-wise approach, the features of
all new points would need to be distributed around the machines.
Second, as we mentioned above, the update direction from the
block-diagonal approximation of the Hessian can be poor if the data is
distributed nonuniformly across machines, and data is more likely to
be distributed evenly across instances than across features.  Thus,
those algorithms focusing on feature-wise split of $X$ are excluded
from our discussion and empirical comparison.

\section{Numerical Experiments}
\label{sec:exp}
We investigate the empirical performance of DPLBFGS for solving both
the primal and dual problems \eqref{eq:primal} and \eqref{eq:dual} on
binary classification problems with training data points $(\bx_i, y_i)
\in \R^d \times \{-1,1\}$ for $i=1,\dotsc,n$. 
For the primal problem, we consider solving $\ell_1$-regularized
logistic regression problems:
\begin{equation}
P(\bw) = C\sum_{i=1}^n \log\left(1 + e^{-y_i \bx_i^\top
	\bw}\right) + \|\bw\|_1,
\label{eq:logistic}
\end{equation}
where $C > 0$ is a parameter prespecified to trade-off between
the loss term and the regularization term.
Note that since the logarithm term is nonnegative, the regularization
term ensures that the level set is bounded.
Therefore, within the bounded set, the loss function is strongly
convex with respect to $X^\top \bw$ and the regularizer can be
reformulated as a polyhedron
constrained linear term. One can thus easily show that
\eqref{eq:logistic} satisfies the quadratic growth condition
\eqref{eq:qg}.
Therefore, our algorithm enjoys global linear convergence on this
problem.

For the dual problem, we consider $\ell_2$-regularized squared-hinge
loss problems, which is of the form
\begin{equation}
	D(\AL) = \frac12 \left\| Y X\AL\right\|_2^2 + \frac{1}{4C} \|\AL\|_2^2
	- \mathbf{1}^\top \AL
	+ {\mathbb{1}}_{\R_+^n}\left(\AL\right),
	\label{eq:l2svm}
\end{equation}
where $Y$ is the diagonal matrix consists of the labels $y_i$,
$\mathbf{1} = (1,\dots,1)$ is the vector of ones,
given a convex set $\mathbf{X}$, ${\mathbb{1}}_{\mathbf{X}}$ is its
indicator function such that
\begin{equation*}
	{\mathbb{1}}_{\mathbf{X}}(x) = \begin{cases}
		0 &\text{ if } x \in X,\\
		\infty &\text{ else},
	\end{cases}
\end{equation*}
and $\R_+^n$ is the nonnegative orthant in $\R^n$.
This strongly convex quadratic problem is considered for easier
implementation of the Catalyst framework in comparison.

We consider the publicly available binary classification data sets
listed in Table~\ref{tbl:data},\footnote{Downloaded from
  \url{https://www.csie.ntu.edu.tw/~cjlin/libsvmtools/datasets/}.}
and partitioned the instances evenly across machines.
$C$ is fixed to $1$ in all our experiments for simplicity.
We ran our experiments on a local cluster of $16$ machines running
MPICH2, and all algorithms are implemented in C/C++.  The inversion of
$M$ defined in \eqref{eq:M} is performed through LAPACK
\citep{And99a}.  The comparison criteria are the relative objective
error
\begin{equation*}
	\left|\frac{F(x) - F^*}{F^*}\right|
\end{equation*}
versus either the amount communicated (divided by $d$) or the overall
running time,
where $F^*$ is the optimal objective value,
and $F$ can be either the primal objective $P(\bw)$ or the dual
objective $D(\AL)$, depending on which problem is being considered.
The former criterion is useful in estimating the performance in
environments in which communication cost is extremely high.

The parameters of our algorithm were set as follows:
$\theta = 0.5$, $\beta = 2$, $\sigma_0 = 10^{-2}$, $\sigma_1 = 10^{-4}$,
$m=10$, $\delta = 10^{-10}$.
The parameters in SpaRSA follow the setting in \cite{WriNF09a},
$\theta$ is set to halve the step size each time, the value of
$\sigma_0$ follows the default experimental setting of
\cite{LeeWCL17a}, $\delta$ is set to a small enough value, and $m=10$
is a common choice for LBFGS.
The code used in our experiments is
available at \url{http://github.com/leepei/dplbfgs/}.

In all experiments, we show results of the backtracking variant only,
as we do not observe significant difference in performance between the
line-search approach and the trust-region approach in our algorithm.

\begin{table}[tb]
\caption{Data statistics.}
\label{tbl:data}
\centering
\begin{tabular}{@{}l|rr|rr}
Data set & $n$ (\#instances) & $d$ (\#features) & \#nonzeros\\
\hline
news & 19,996 & 1,355,191 & 9,097,916 \\
epsilon & 400,000 & 2,000 & 800,000,000\\
webspam & 350,000 & 16,609,143 & 1,304,697,446\\
avazu-site & 25,832,830 & 999,962 & 387,492,144
\end{tabular}
\end{table}

In the subsequent experiments,
we first use the primal problem \eqref{eq:logistic} to examine how
inexactness of the subproblem solution affects the communication
complexity, overall running time, and step sizes.
We then compare our algorithm with state of the art distributed
solvers for \eqref{eq:logistic}.
Finally, comparison on the dual problem \eqref{eq:l2svm} is conducted.

\subsection{Effect of Inexactness in the Subproblem Solution}
We first examine how the degree of inexactness of the approximate
solution of subproblems \eqref{eq:quadratic} affects the convergence
of the overall algorithm. Instead of treating SpaRSA as a steadily
linearly converging algorithm, we take it as an algorithm that
sometimes decreases the objective much faster than the worst-case
guarantee, thus an adaptive stopping condition is used.  In
particular, we terminate Algorithm \ref{alg:sparsa} when the norm of
the current update step is smaller than $\epsilon_1$ times that of the
first update step, for some prespecified $\epsilon_1 > 0$.  From the
proof of Lemma~\ref{lemma:sparsa}, the norm of the update step bounds
the value of $Q(p) - Q^*$ both from above and from below (assuming
exact solution of \eqref{eq:dk}, which is indeed the case for
the selected problems), and thus serves as a good measure of the
solution precision.  In Table~\ref{tbl:stop}, we compare runs with the
values $\epsilon_1 =
10^{-1}, 10^{-2}, 10^{-3}$. For the datasets news20 and webspam, it is
as expected that tighter solution of \eqref{eq:quadratic} results in
better updates and hence lower communication cost,
though it may not result in a shorter convergence time because of more
computation per round.
As for the dataset epsilon, which has a
smaller data dimension $d$, the $O(m)$ communication cost per SpaRSA
iteration for calculating $\nabla \hat f$ is significant in
comparison. In this case, setting a tighter stopping criterion for
SpaRSA can incur higher communication cost and longer running
time.

In Table~\ref{tbl:steps}, we show the distribution of the step sizes
over the main iterations, for the same set of values of $\epsilon_1$.  As
we discussed in Section~\ref{sec:analysis}, although the smallest
$\lambda$ can be much smaller than one, the unit step is usually
accepted.  Therefore, although the worst-case communication complexity
analysis is dominated by the smallest step encountered, the practical
behavior is much better.
This result also suggests that the difference between DPLBFGS-LS and
DPLBFGS-TR should be negligible, as most of the times, the original
$H$ with unit step size is accepted.

\begin{table}[tb]
\caption{Different stopping conditions of SpaRSA as an approximate
solver for \eqref{eq:quadratic}. We show required amount of
communication (divided by $d$) and running time (in seconds) to reach
$F(\bw) - F^* \leq 10^{-3} F^*$.}
\label{tbl:stop}
\centering
	\begin{tabular}{l|r|r|r}
Data set & $\epsilon_1$ & Communication & Time\\
\hline
\multirow{3}{*}{news20} & $10^{-1}$ & 28 & 11 \\
& $10^{-2}$ & 25 & 11 \\
& $10^{-3}$ & 23 & 14 \\
\hline
\multirow{3}{*}{epsilon} & $10^{-1}$ & 144 & 45 \\
& $10^{-2}$ & 357 & 61 \\
& $10^{-3}$ & 687 & 60 \\
\hline
\multirow{3}{*}{webspam} & $10^{-1}$ & 452 & 3254 \\
& $10^{-2}$ & 273 & 1814 \\
& $10^{-3}$ & 249 & 1419
\end{tabular}
\end{table}
\begin{table}[tb]
\caption{Step size distributions.}
\label{tbl:steps}
\centering
\begin{tabular}{l|r|r|r}
Data set & $\epsilon_1$ & percent of $\lambda = 1$ & smallest $\lambda$\\
\hline
\multirow{3}{*}{news20} & $10^{-1}$ & $95.5\%$ & $2^{-3}$ \\
& $10^{-2}$ & $95.5\%$ & $2^{-4}$ \\
& $10^{-3}$ & $95.5\%$ & $2^{-3}$ \\
\hline
\multirow{3}{*}{epsilon} & $10^{-1}$ & $96.8\%$ & $2^{-5}$ \\
& $10^{-2}$ & $93.4\%$ & $2^{-6}$ \\
& $10^{-3}$ & $91.2\%$ & $2^{-3}$ \\
\hline
\multirow{3}{*}{webspam} & $10^{-1}$ & $98.5\%$ & $2^{-3}$ \\
& $10^{-2}$ & $97.6\%$ & $2^{-2}$ \\
& $10^{-3}$ & $97.2\%$ & $2^{-2}$
\end{tabular}
\end{table}

\subsection{Comparison with Other Methods for the Primal Problem}
Now we compare our method with two state-of-the-art distributed
algorithms for \eqref{eq:f}.  In addition to a proximal-gradient-type
method that can be used to solve general \eqref{eq:f} in distributed
environments easily, we also include one solver specifically designed
for $\ell_1$-regularized problems in our comparison.  These methods
are:
\begin{itemize}
	\item DPLBFGS-LS: our Distributed Proximal LBFGS approach. We fix
		$\epsilon_1 = 10^{-2}$.
	\item SpaRSA \citep{WriNF09a}: the method described in
		Section~\ref{subsec:sparsa}, but applied directly to
		\eqref{eq:primal} but not to the subproblem
		\eqref{eq:quadratic}.
	\item OWLQN \citep{AndG07a}: an orthant-wise quasi-Newton method
		specifically designed for $\ell_1$-regularized problems.
		We fix $m=10$ in the LBFGS approximation.
\end{itemize}
All methods are implemented in C/C++ and MPI.
As OWLQN does not update the coordinates $i$ such that
$-X_{i,:}\nabla \xi(X^T \bw) \in \partial g_i(\bw_i)$ given any $\bw$,
the same preliminary active set selection is applied to our algorithm
to reduce the subproblem dimension and the computational cost, but
note that this does not reduce the communication cost as the gradient
calculation still requires communication of a full $d$-dimensional vector.

The AG method \citep{Nes13a} can be an alternative to
SpaRSA, but its empirical performance has been shown to be similar to
SpaRSA \citep{YanZ11a} and it requires strong convexity and Lipschitz
parameters to be estimated, which induces an additional cost.

A further examination on different values of $m$ indicates that
convergence speed of our method improves with larger $m$,
while in OWLQN, larger $m$  usually does not lead to better results.
We use the same value of $m$ for both methods and choose a value that
favors OWLQN.

\begin{figure}
\centering
\begin{tabular}{cc}
	Communication & Time\\
	\includegraphics[width=.40\linewidth]{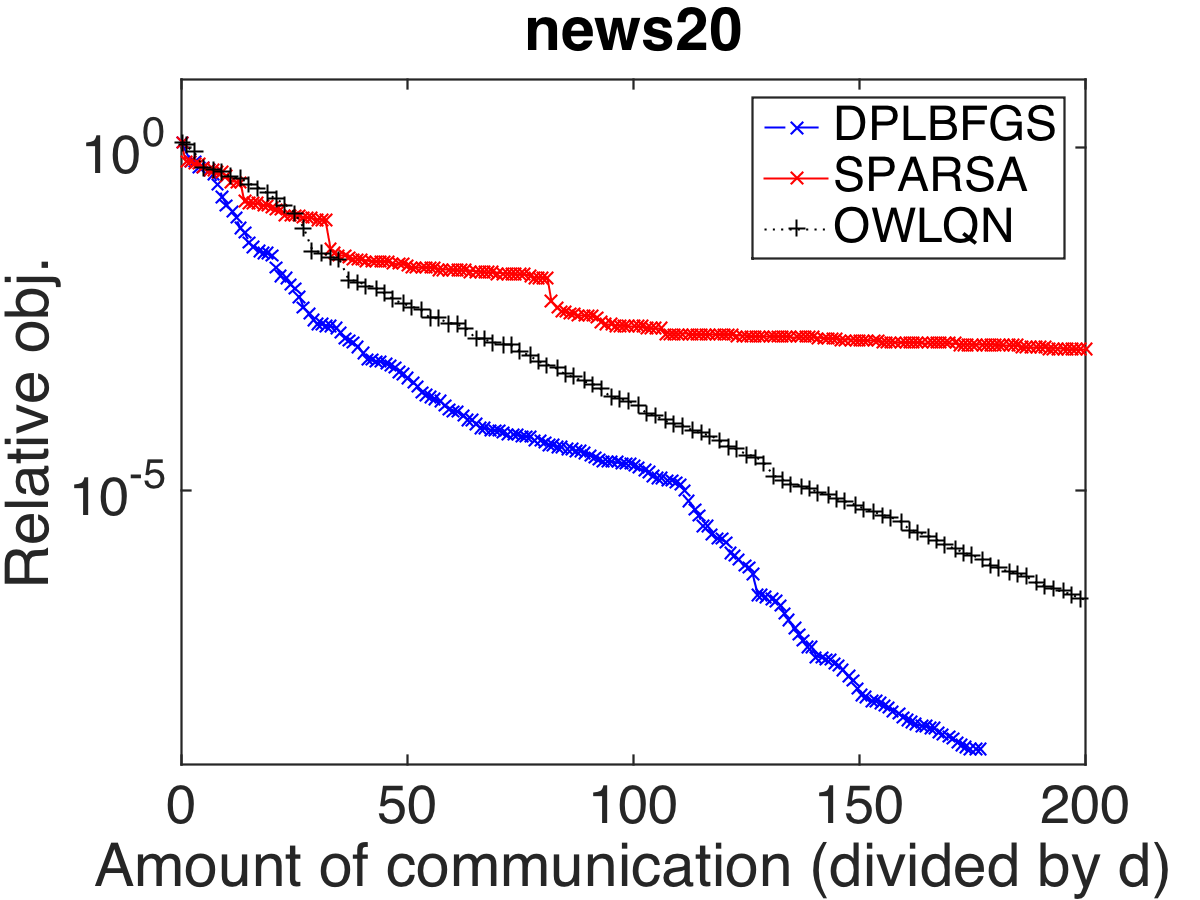}&
	\includegraphics[width=.40\linewidth]{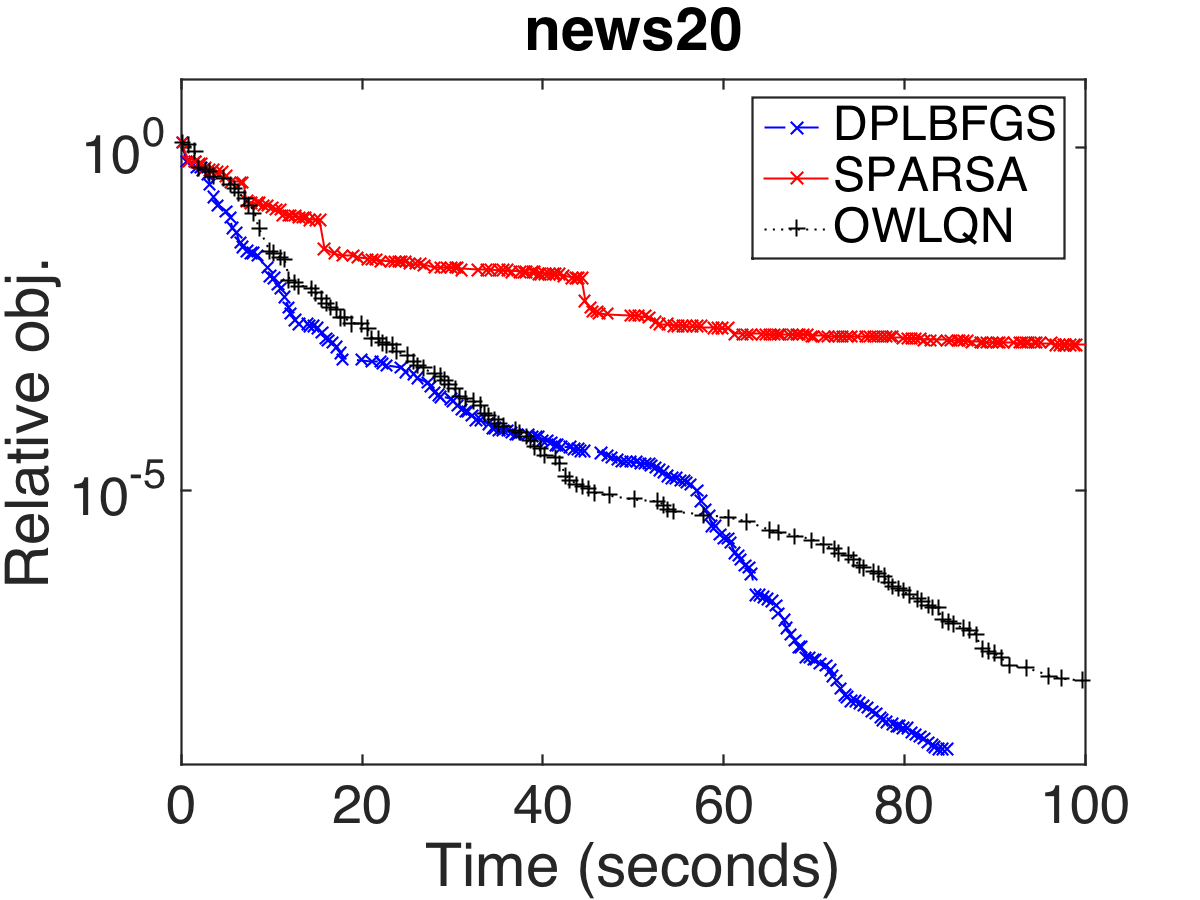}\\
	\includegraphics[width=.40\linewidth]{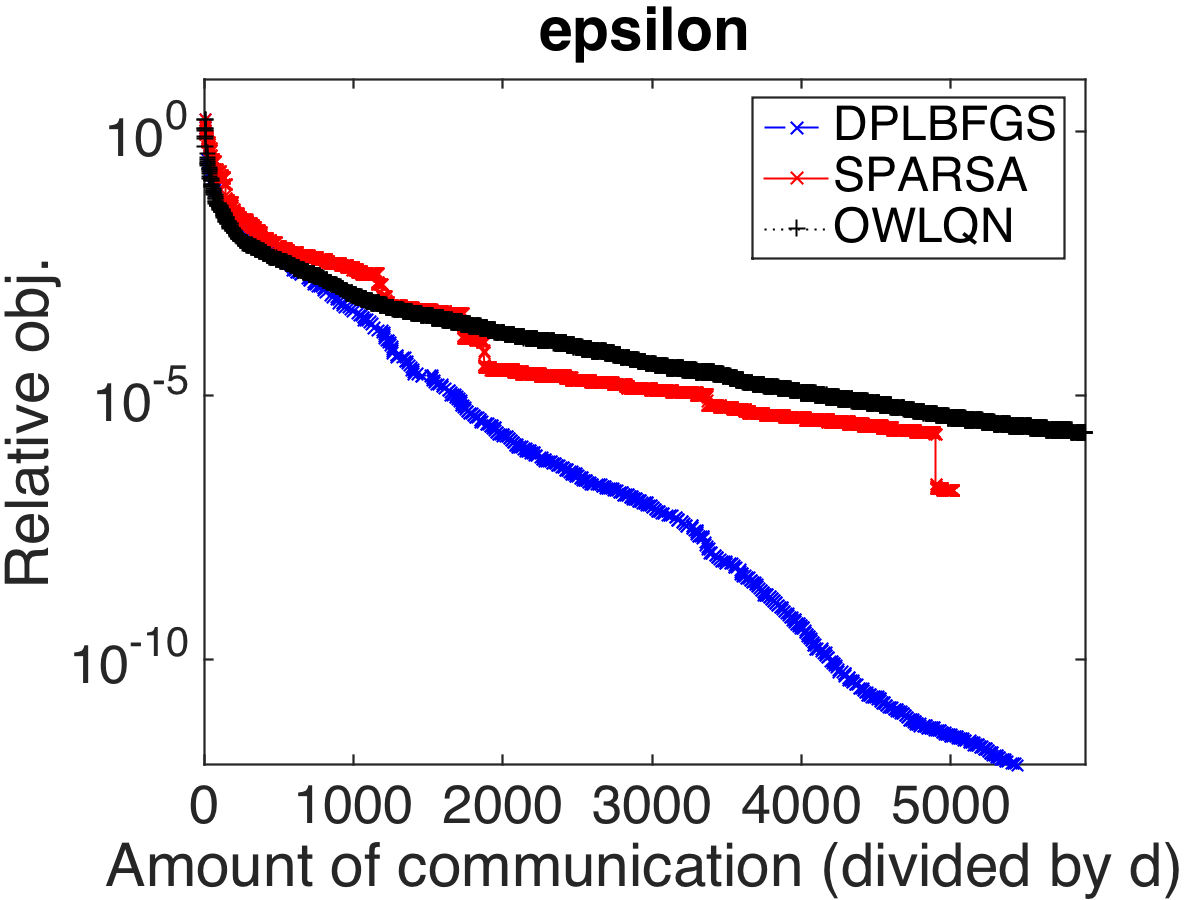}&
	\includegraphics[width=.40\linewidth]{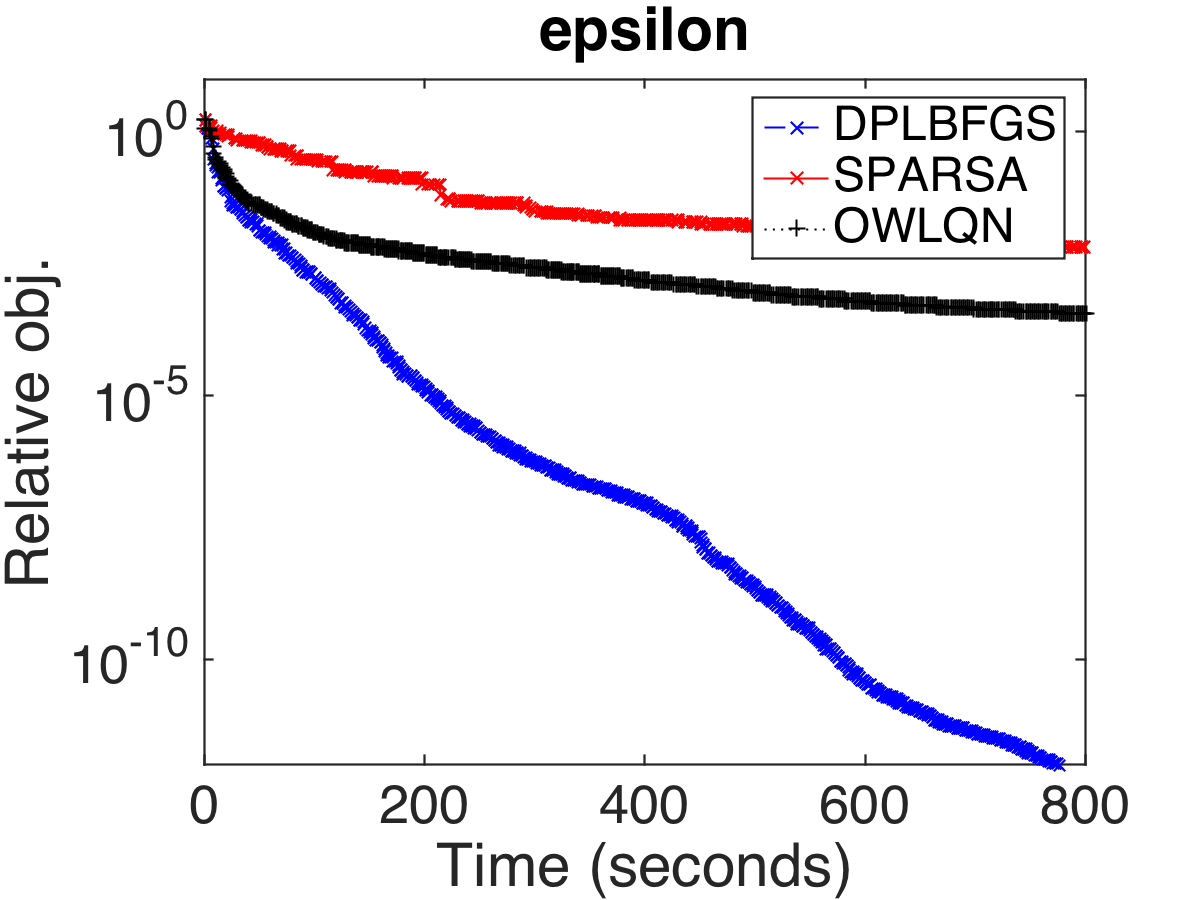}\\
	\includegraphics[width=.40\linewidth]{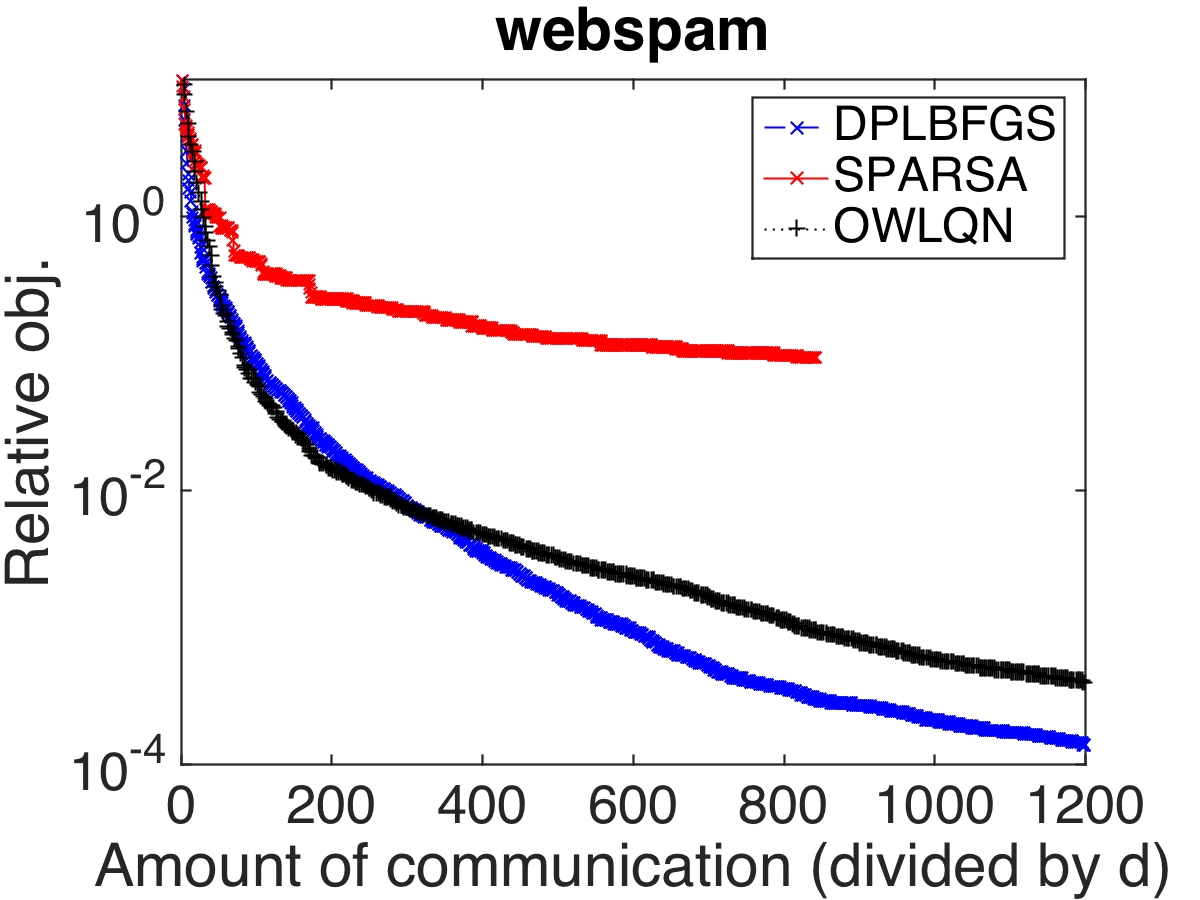}&
	\includegraphics[width=.40\linewidth]{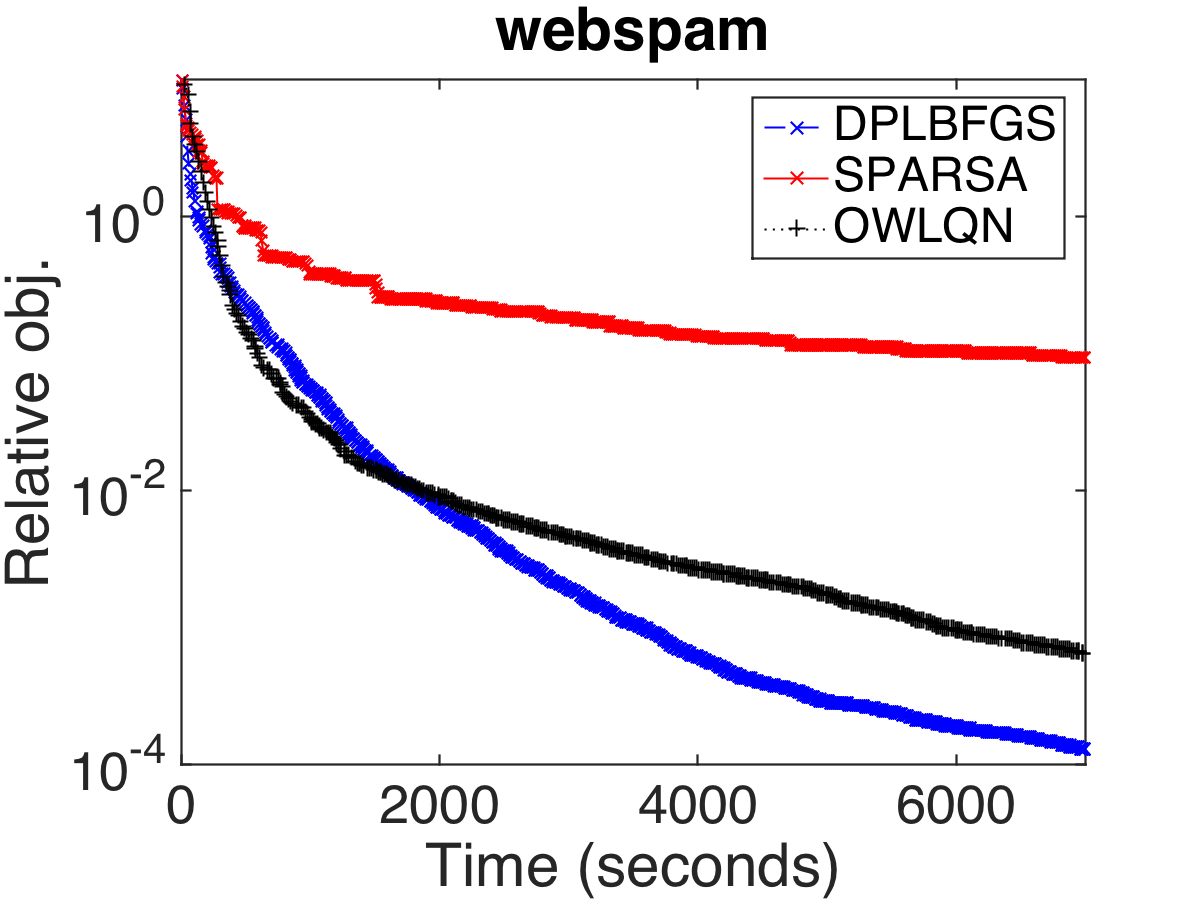}\\
	\includegraphics[width=.40\linewidth]{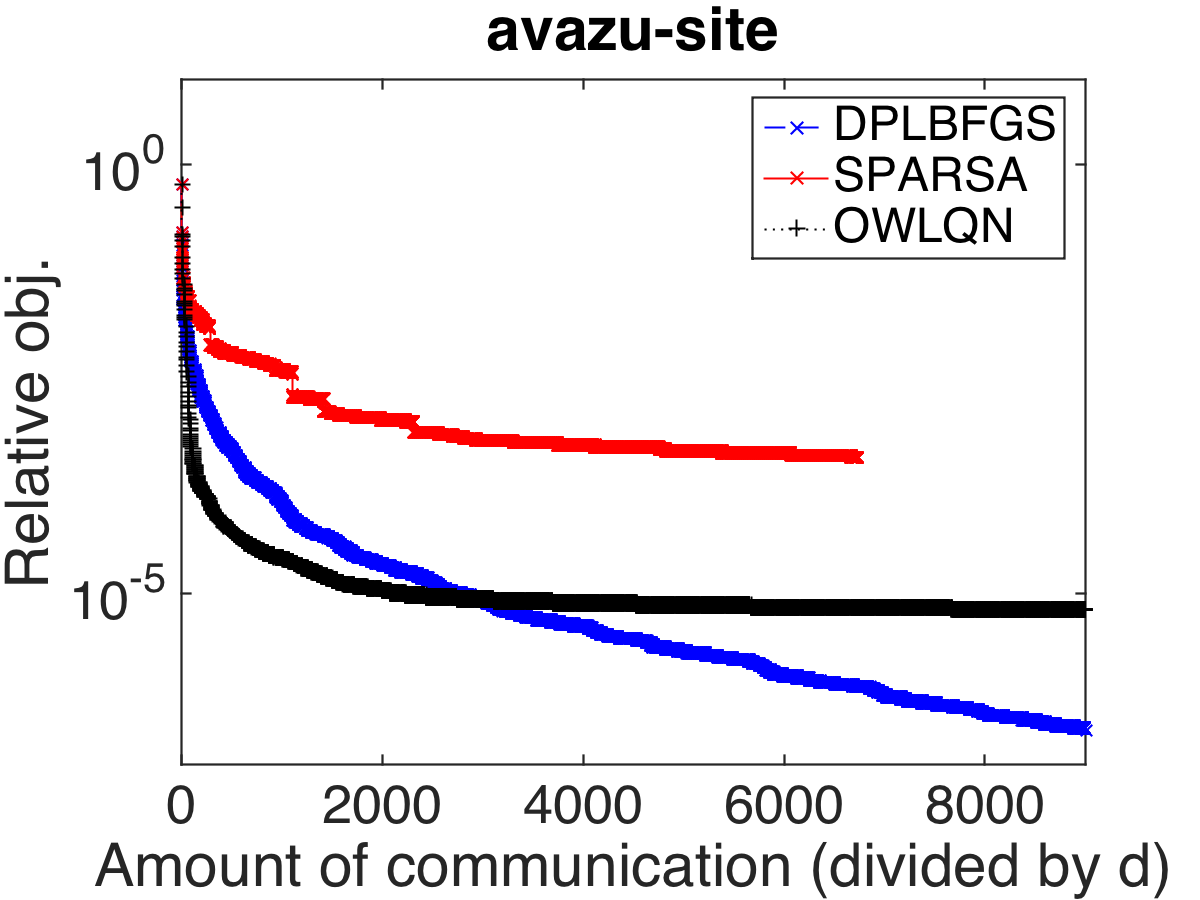}&
	\includegraphics[width=.40\linewidth]{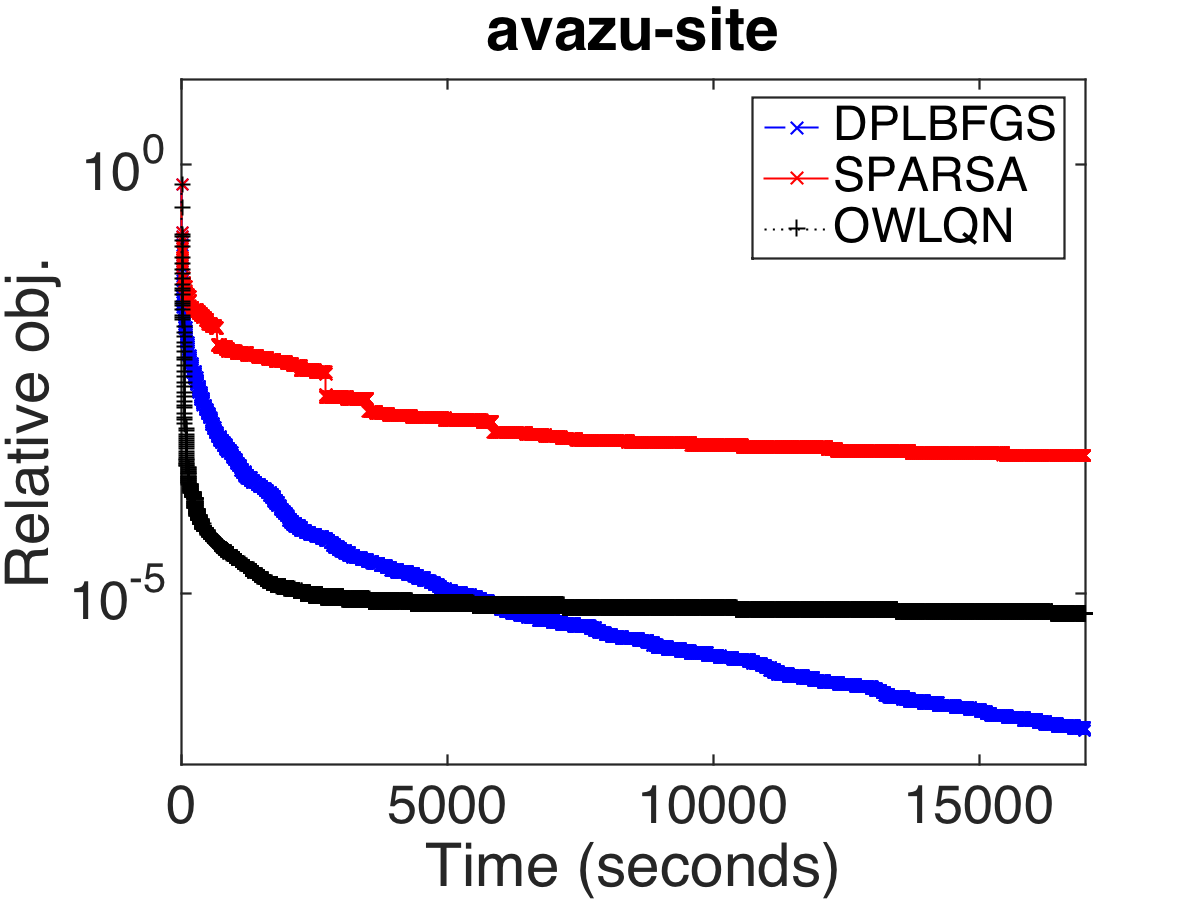}
\end{tabular}
	\caption{Comparison between different methods for
		\eqref{eq:logistic} in terms of relative objective difference
		to the optimum. Left: communication
	(divided by $d$); right: running time (in seconds).}
\label{fig:compare}
\end{figure}

The results are provided in Figure~\ref{fig:compare}.
Our method is always the fastest in both criteria.  For
epsilon, our method is orders of magnitude faster, showing that
correctly using the curvature information of the smooth part is indeed
beneficial in reducing the communication complexity.

It is possible to include specific heuristics for
$\ell_1$-regularized problems, such as those applied in
\citet{YuaHL12a, KaiYDR14a}, to further accelerate our method for this
problem, and the exploration on this direction is an interesting topic
for future work.

\subsection{Comparison on the Dual Problem}
\label{subsec:dualexp}
\begin{figure}[tb]
\centering
\begin{tabular}{cc}
	Communication & Time\\
	\includegraphics[width=.40\linewidth]{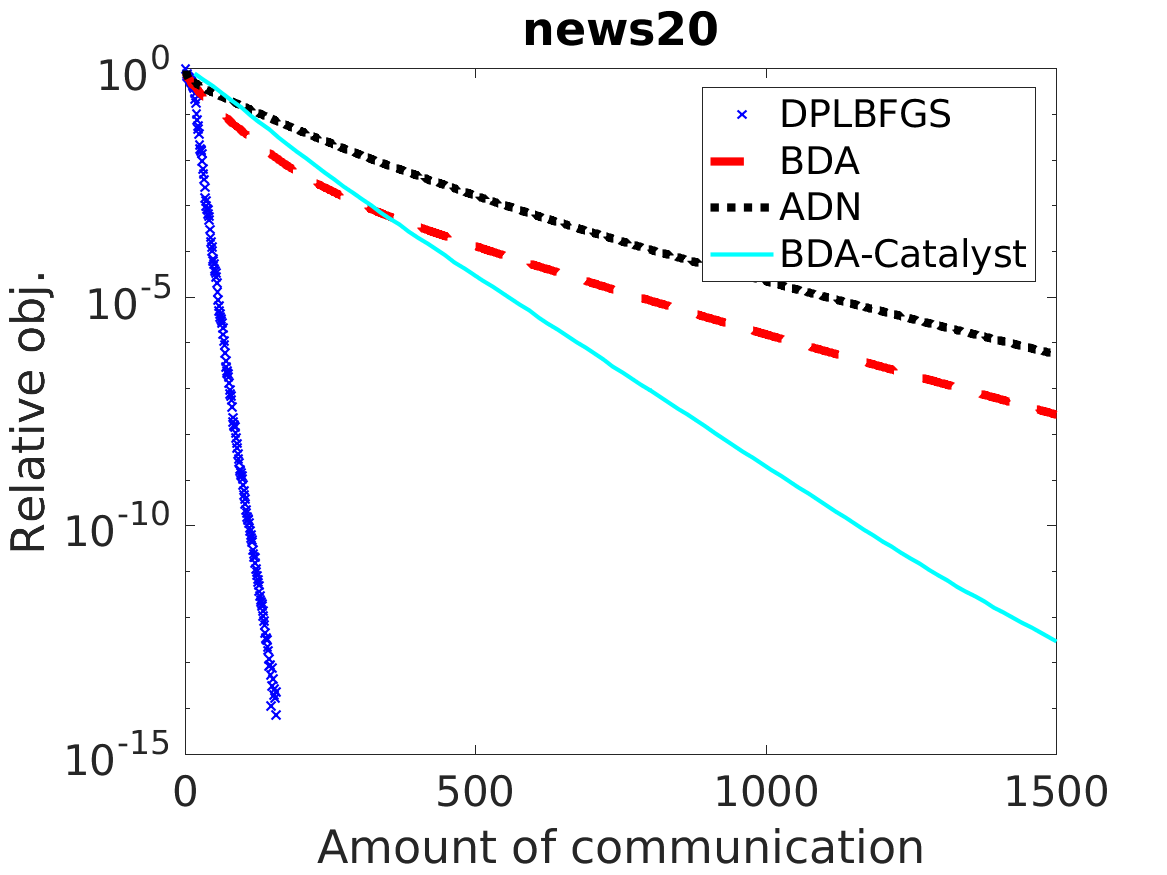}&
	\includegraphics[width=.40\linewidth]{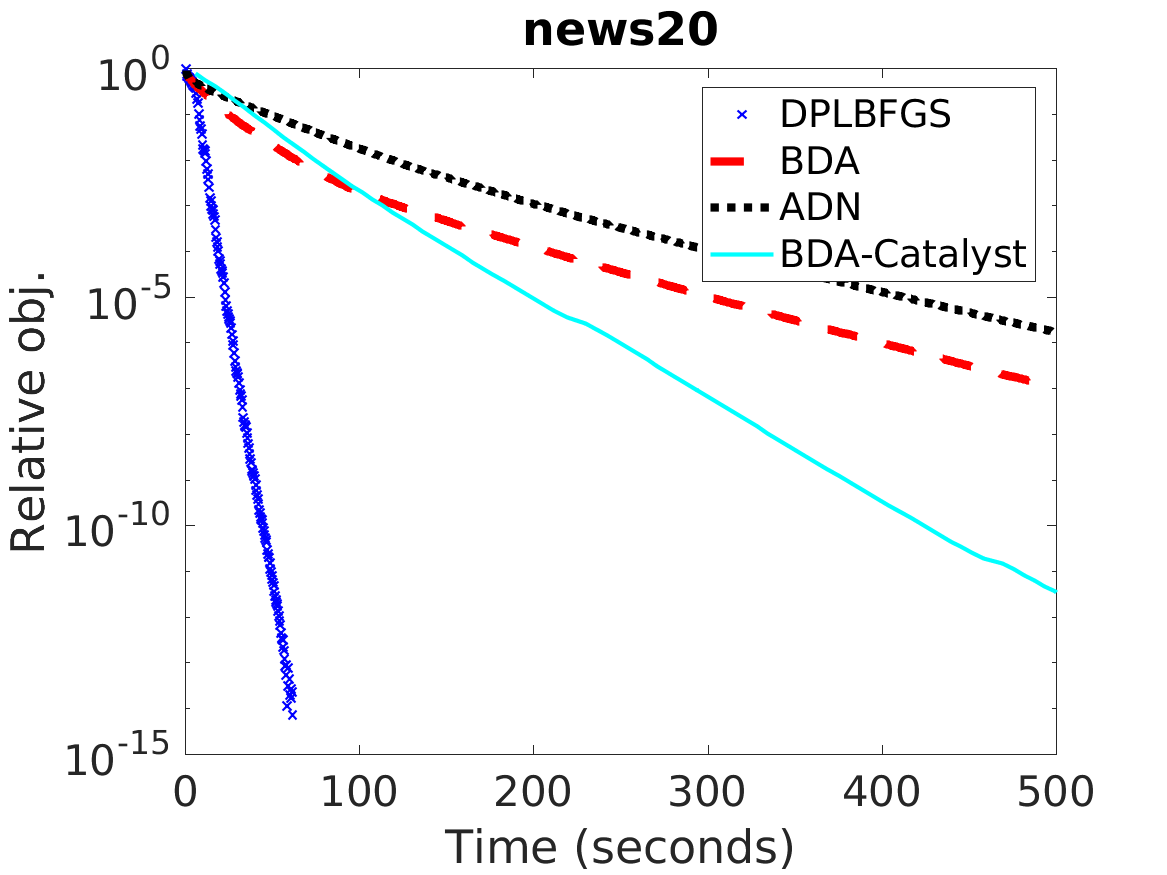}\\
	\includegraphics[width=.40\linewidth]{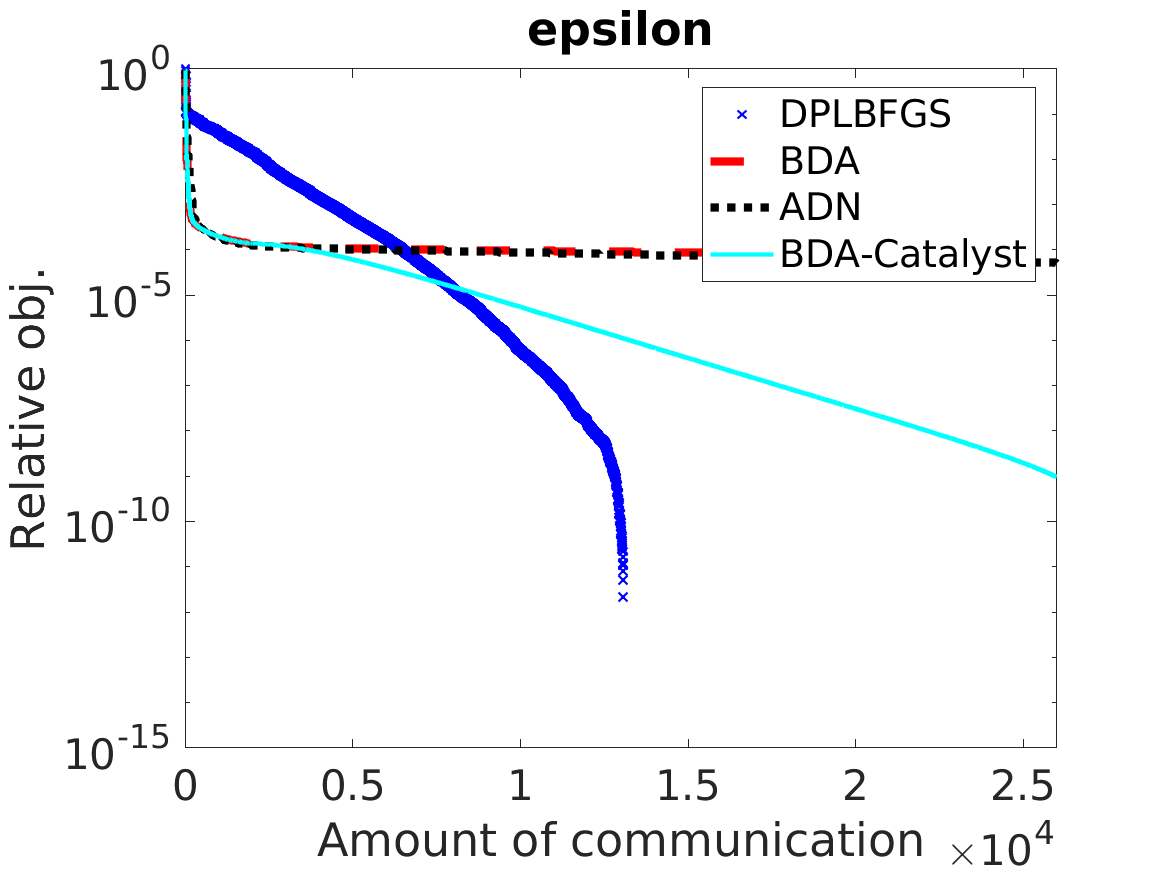}&
	\includegraphics[width=.40\linewidth]{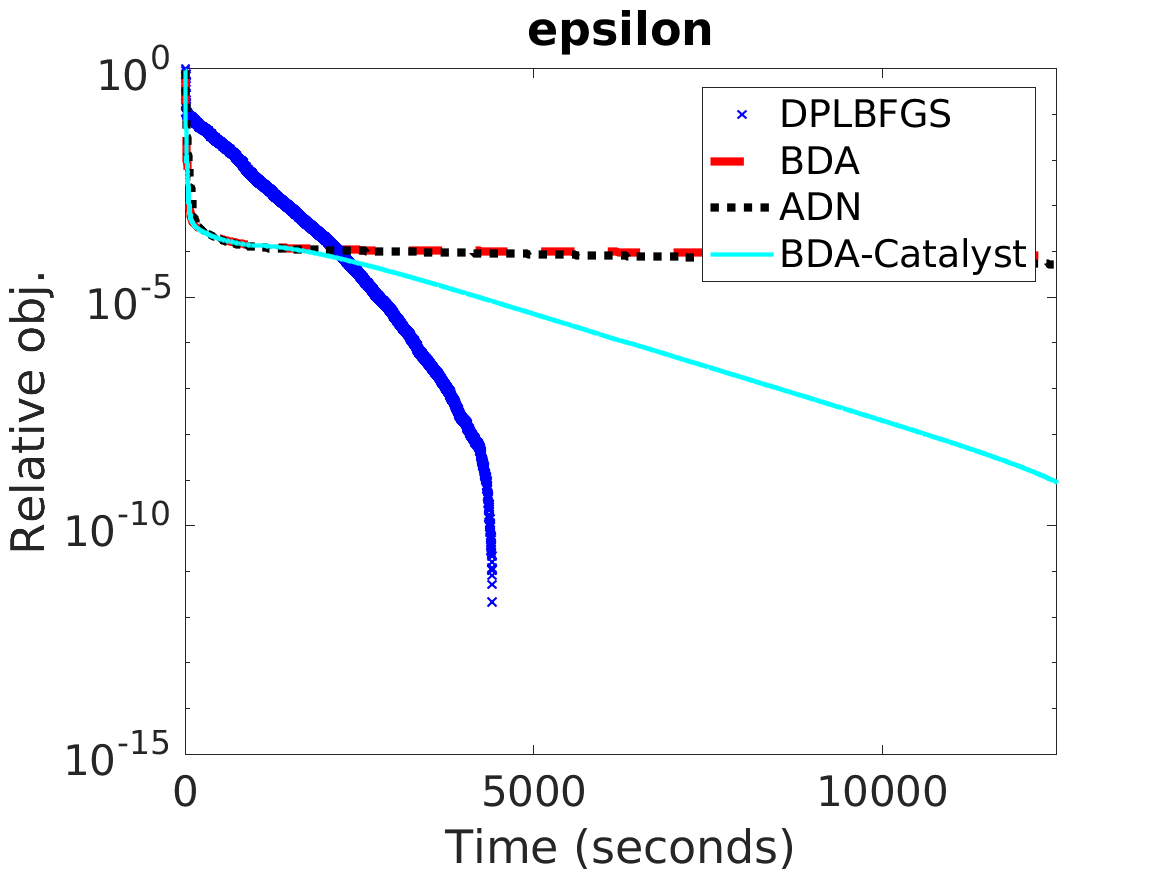}\\
	\includegraphics[width=.40\linewidth]{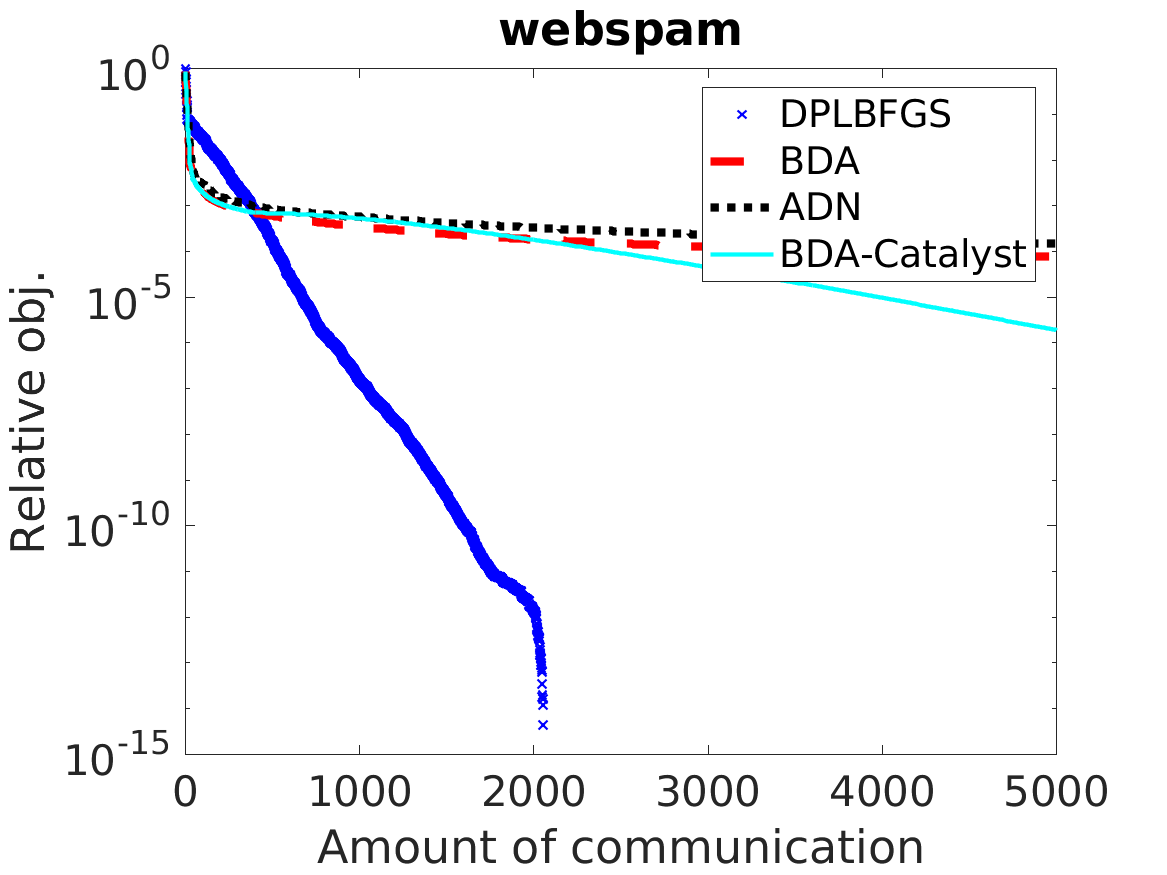}&
	\includegraphics[width=.40\linewidth]{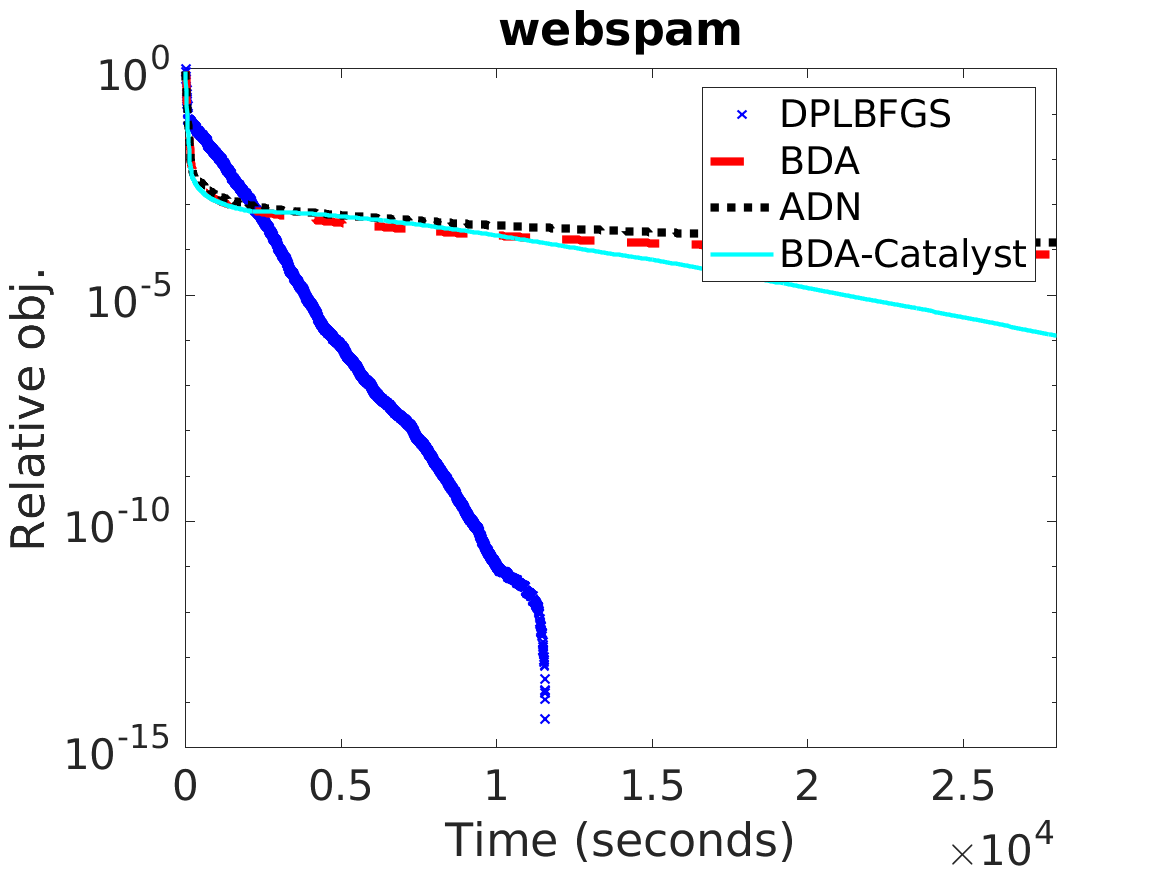}
\end{tabular}
\caption{Comparison between different methods for \eqref{eq:l2svm} in
terms of relative objective difference to the optimum. Left:
communication (divided by $d$); right: running time (in seconds).}
\label{fig:dual}
\end{figure}

\begin{figure}[tb]
\centering
\begin{tabular}{cc}
	Communication & Time\\
	\includegraphics[width=.40\linewidth]{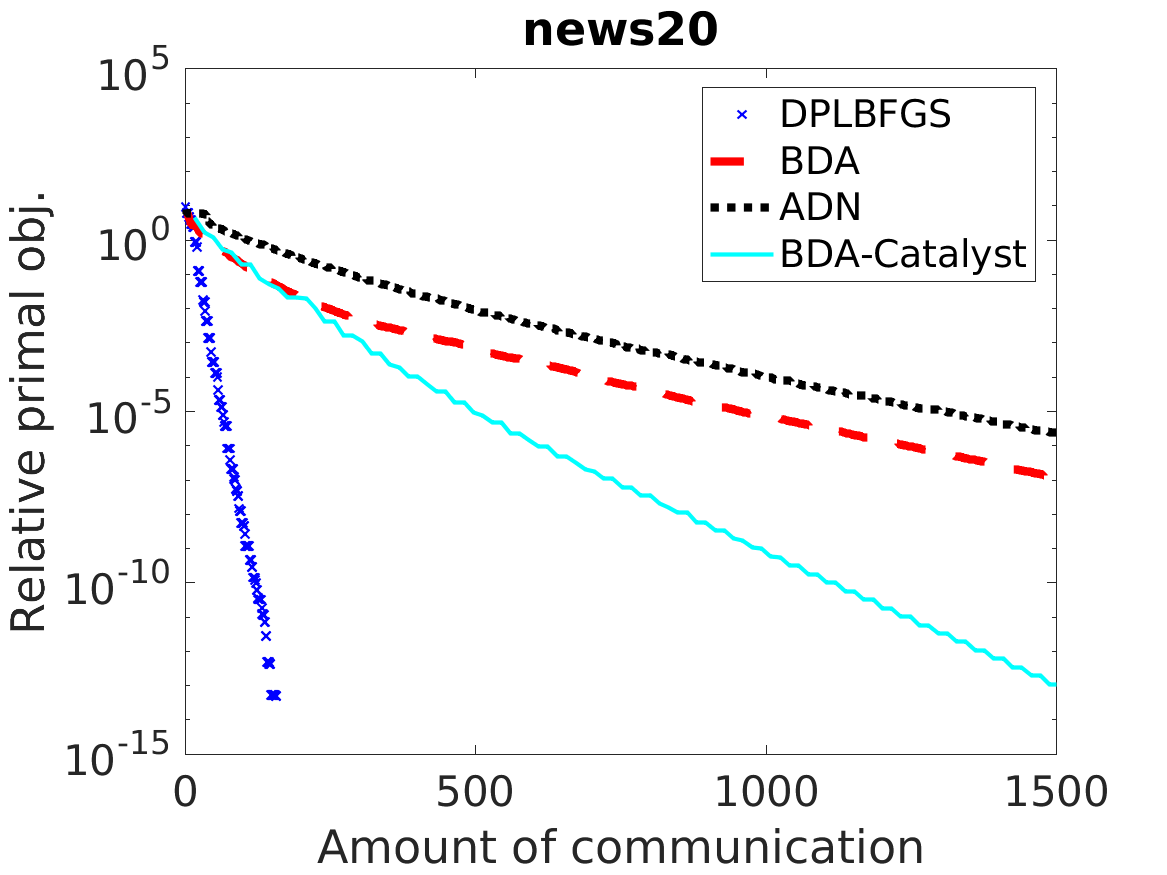}&
	\includegraphics[width=.40\linewidth]{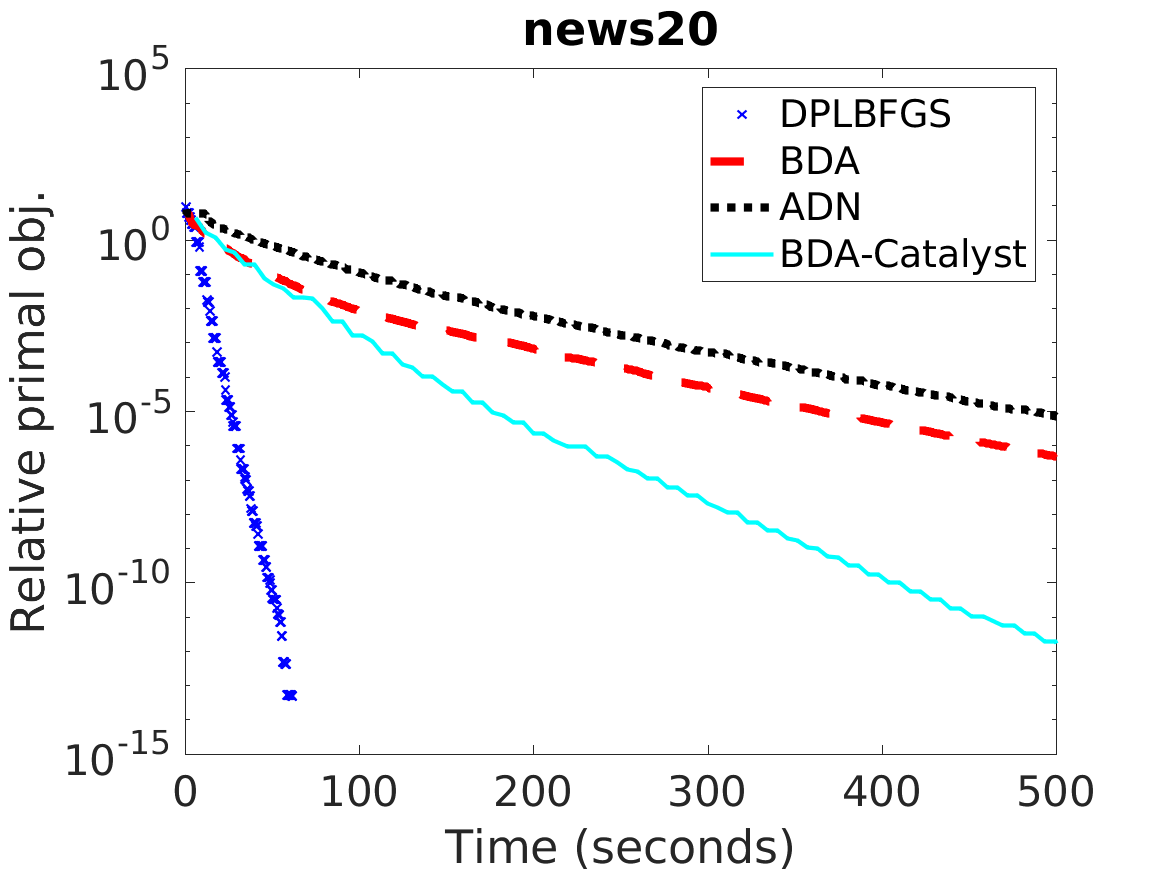}\\
	\includegraphics[width=.40\linewidth]{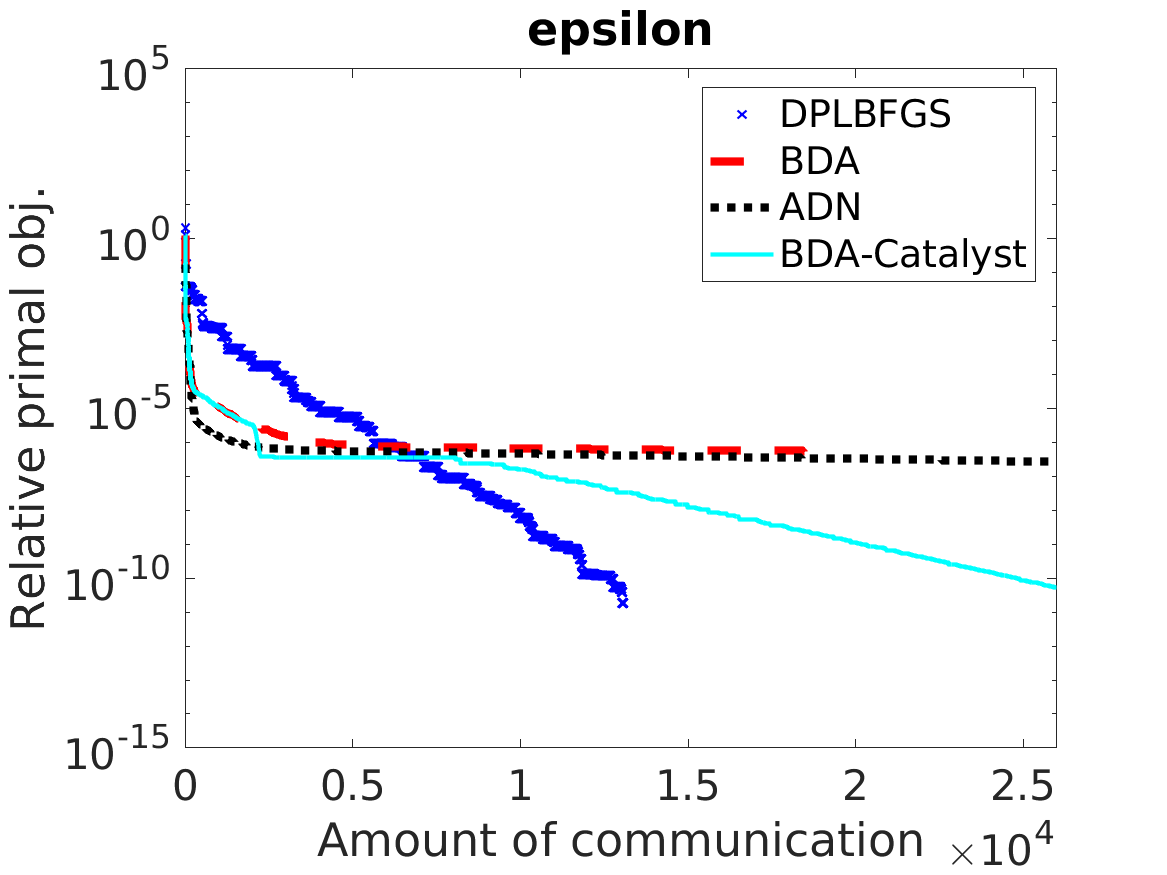}&
	\includegraphics[width=.40\linewidth]{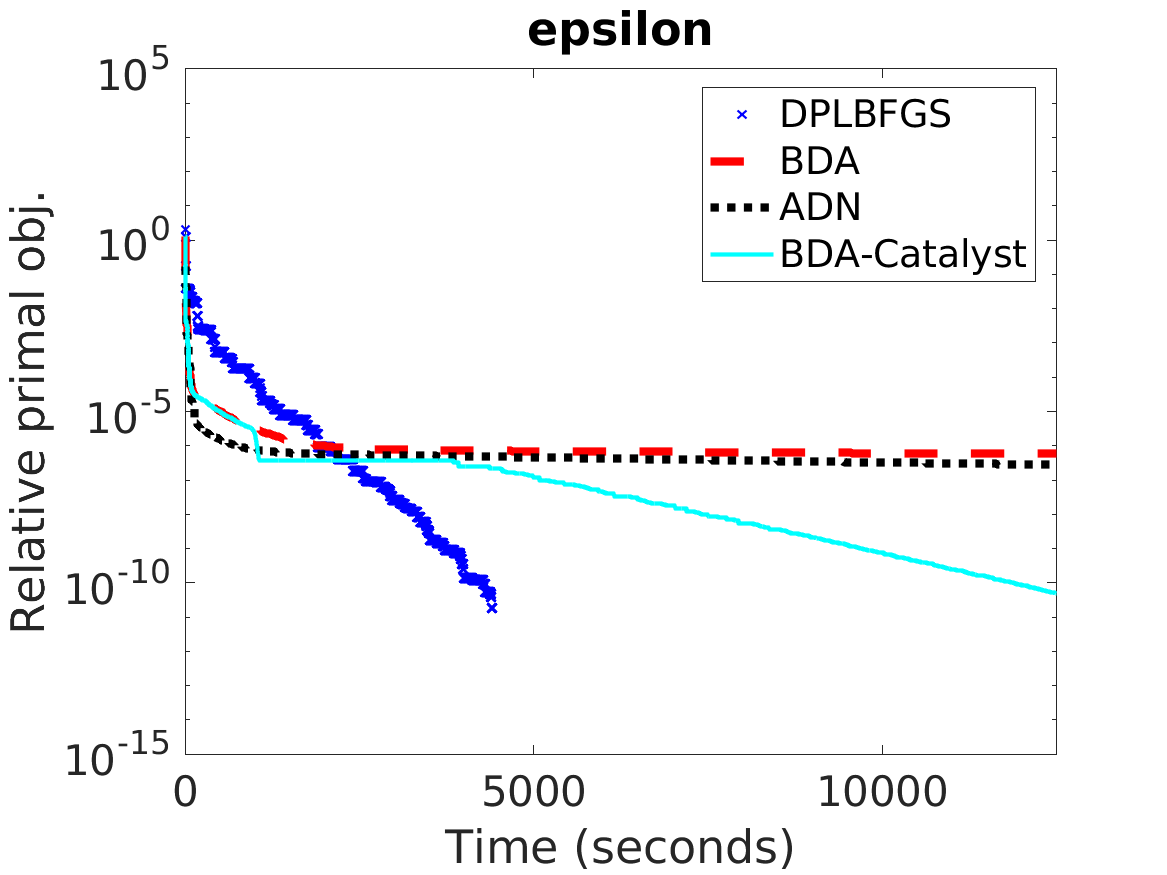}\\
	\includegraphics[width=.40\linewidth]{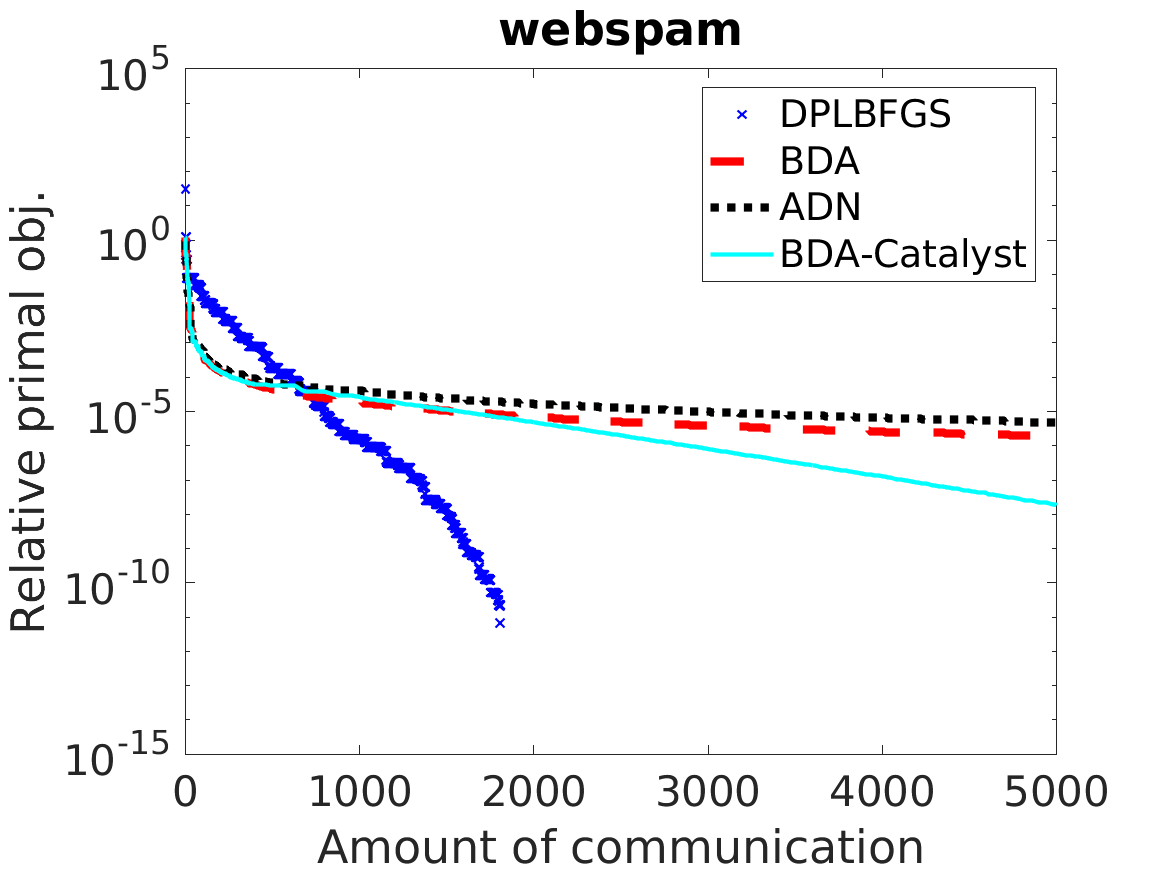}&
	\includegraphics[width=.40\linewidth]{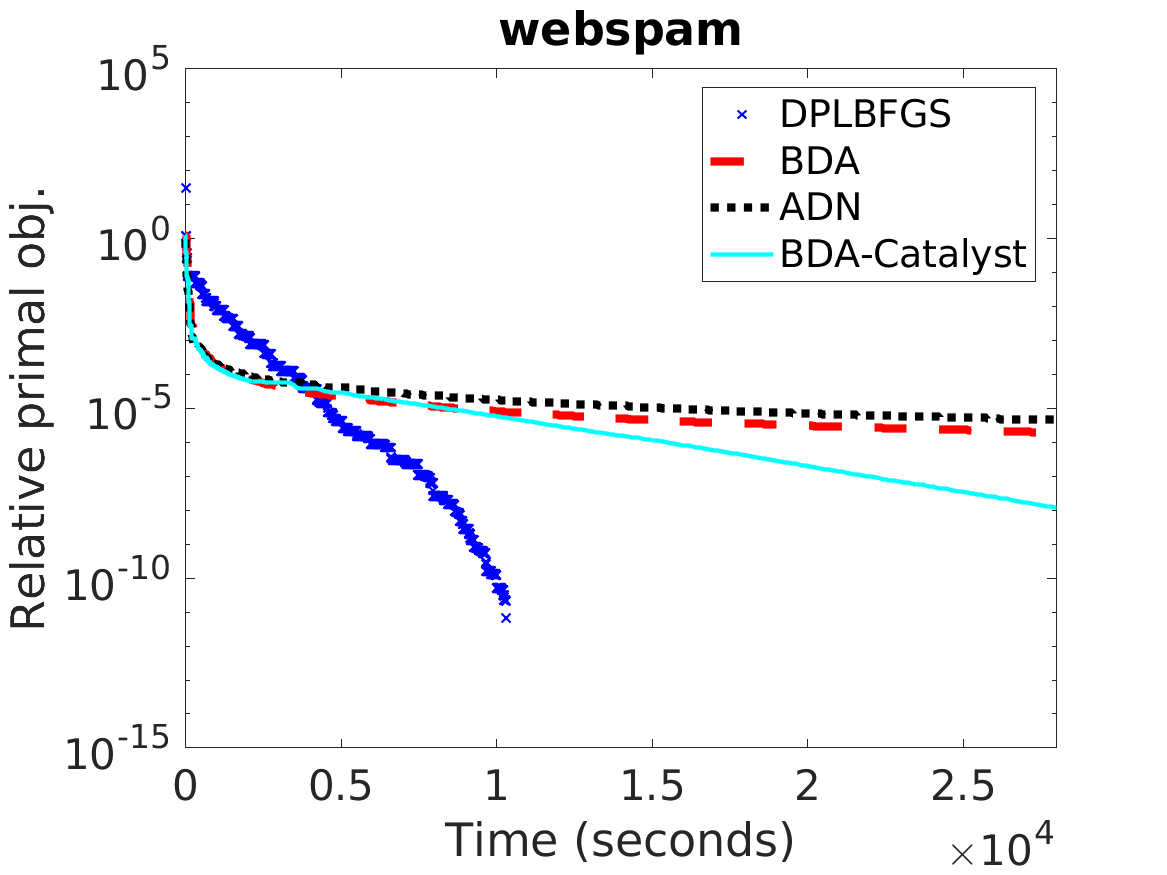}
\end{tabular}
\caption{Comparison between different methods for \eqref{eq:l2svm} in
terms of relative \textit{primal} objective difference to the optimum. Left:
communication (divided by $d$); right: running time (in seconds).}
\label{fig:dual-primal}
\end{figure}

Now we turn to solve the dual problem, considering the specific
example \eqref{eq:l2svm}.
We compare the following algorithms.
\begin{itemize}
	\item BDA \citep{LeeC17a}: a distributed algorithm using
		Block-Diagonal Approximation of the real Hessian of the smooth
		part with line search.
	\item BDA with Catalyst: using the BDA algorithm
		within the Catalyst framework \citep{LinMH15a} for accelerating
		first-order methods.
	\item ADN \citep{DunLGBHJ18a}: a trust-region approach where
		the quadratic term is a multiple of the block-diagonal part of
		the Hessian, scaled adaptively as the algorithm progresses.
	\item DPLBFGS-LS: our Distributed Proximal LBFGS approach. We fix
		$\epsilon_1 = 10^{-2}$ and limit the number of SpaRSA
		iterations to $100$. For the first ten iterations when
		$m(t) < m$, we use BDA to generate the update steps instead.
\end{itemize}
For BDA, we use the C/C++ implementation in the package
MPI-LIBLINEAR.\footnote{\url{http://www.csie.ntu.edu.tw/~cjlin/libsvmtools/distributed-liblinear/}.}
We implement ADN by modifying the above implementation of BDA.
In both BDA and ADN, following \cite{LeeC17a} we use random-permutation
coordinate descent (RPCD) for the local subproblems, and for each outer
iteration we perform one epoch of RPCD.
For the line search step in both BDA and DPLBFGS-LS, since the
objective \eqref{eq:l2svm} is quadratic, we can find the exact minimizer
efficiently (in closed form). The convergence guarantees still holds
for exact line search, so we use this here in place of the backtracking
approach described earlier.

We also applied the Catalyst framework \citep{LinMH15a} for
accelerating first-order methods to BDA to tackle the dual problem,
especially for dealing with the stagnant convergence issue.
This framework requires a good estimate of the convergence rate and
the strong convexity parameter $\sigma$.
From \eqref{eq:l2svm}, we know that $\sigma = 1 / (2C)$,
but the actual convergence rate is hard to estimate as BDA
interpolates between (stochastic) proximal coordinate descent (when
only one machine is used) and proximal gradient (when $n$ machines are
used).
After experimenting with different sets of parameters for BDA with
Catalyst, we found the following to work most effectively: for
every outer iteration of the Catalyst framework, $K$ iterations of BDA
is conducted with early termination if a negative step size is obtained
from exact line search;
for the next Catalyst iteration, the warm-start initial point is
simply the iterate at the end of the previous Catalyst iteration;
before starting Catalyst, we run the unaccelerated version of BDA for
certain iterations to utilize its advantage of fast early convergence.
Unfortunately, we do not find a good way to estimate the $\kappa$ term
in the Catalyst framework that works for all data sets. Therefore, we
find the best $\kappa$ by a grid search.
We provide a detailed description of our implementation of the
Catalyst framework on this problem and the related parameters used in
this experiment in Appendix \ref{app:catalyst}.

We focus on the combination of Catalyst and BDA (instead of with ADN)
for a few reasons. Since both BDA and ADN are distributed
methods that use the block-diagonal portion of the Hessian matrix, it
should suffice to evaluate the application of Catalyst to the better
performing of the two to represent this class of algorithms. In
addition, dealing with the trust-region adjustment of ADN becomes
complicated as the problem changes through the Catalyst iterations.

The results are shown in Figure \ref{fig:dual}.
We do not present results on the avazu data set in this experiment as
all methods take extremely long time to converge.
We first observe that, contrary to what is claimed in
\cite{DunLGBHJ18a}, BDA outperforms ADN on news20 and webspam, though the
difference is insignificant, and the two are competitive on epsilon.
This also justifies that applying the Catalyst framework on BDA alone
suffices.
Comparing our DPLBFGS approach to the block-diagonal ones, it is clear
that our method performs magnitudes better than the state of the
art in terms of both communication cost and time.  For webspam and
epsilon, the block-diagonal approaches are faster at first, but the
progress stalls after a certain accuracy level.  In contrast, while
the proposed DPLBFGS approach does not converge as rapidly initially,
the algorithm consistently makes progress towards a high accuracy
solution.

As the purpose of solving the dual problem is to obtain an approximate
solution to the primal problem through the formulation
\eqref{eq:walpha}, we are interested on how the methods compare in
terms of the primal solution precision.
This comparison is presented in Figure \ref{fig:dual-primal}.
Since these dual methods are not descent methods for the primal
problem, we apply the pocket approach \citep{Gal90a} suggested in
\cite{LeeC17a} to use the iterate with the smallest primal objective
so far as the current primal solution.
We see that the primal objective values have trends very similar to
the dual counterparts, showing that our DPLBFGS method is also
superior at generating better primal solutions.

A potentially more effective approach is a hybrid one that first uses a
block-diagonal method and then switches over to our DPLBFGS approach after
the block-diagonal method hits the slow convergence phase. Developing
such an algorithm would require a way to determine when we reach such
a stage, and we leave the development of this method to future work.
Another possibility is to consider a structured quasi-Newton approach
to construct a Hessian approximation only for the off-block-diagonal
part so that the block-diagonal part can be utilized simultaneously.

We also remark that our algorithm is partition-invariant in terms of
convergence and communication cost, while the convergence behavior of
the block-diagonal approaches depend heavily on the partition.
This means when more machines are used, these block-diagonal
approaches suffer from poorer convergence, while our method
retains the same efficiency regardless of the number of machines begin
used and how the data points are distributed (except for the
initialization part).

\section{Conclusions} \label{sec:conclusions}
In this work, we propose a practical and communication-efficient
distributed algorithm for solving general regularized nonsmooth ERM
problems.
The proposed approach is the first one that can be applied both to the
primal and the dual ERM problem under the instance-wise split scheme.
Our algorithm enjoys fast performance both theoretically
and empirically and can be applied to a wide range of ERM problems.
Future work for the primal problem include active set identification
for reducing the size of the vector communicated when the solution
exhibits sparsity, and application to nonconvex applications; while
for the dual problem, it is interesting to further exploit the
structure so that the quasi-Newton approach can be combined with real
Hessian entries at the block-diagonal part to get better convergence.


\bibliography{../../inexactprox,../distributederm}

\appendix
\section{Proofs}
In this appendix, we provide proof for Lemma~\ref{lemma:sparsa}.
The rest of Section~\ref{sec:analysis} directly follows the results in
\citet{LeeW18a,PenZZ18a}, and are therefore omitted.
Note that \eqref{eq:strong} implies \eqref{eq:qg}, and
\eqref{eq:qg} implies \eqref{eq:R0} because
$R_0^2$ is upper-bounded by $2(F(x^0) - F^*)/\mu$. Therefore, we get
improved communication complexity by the fast early linear
convergence from the general convex case.
\begin{proof}[Lemma~\ref{lemma:sparsa}]
We prove the three results separately.
\begin{compactenum}
\item
We assume without loss of simplicity that \eqref{eq:safeguard} is
satisfied by all iterations.  When it is not the case, we just need to
shift the indices but the proof remains the same as the pairs of
$(\bs_t, \by_t)$ that do not satisfy \eqref{eq:safeguard} are
discarded.

We first bound $\gamma_t$ defined in \eqref{eq:M}.
From Lipschitz continuity of $\nabla f$, we have that for all $t$,
\begin{equation}
\frac{\|\by_t\|^2}{\by_t^\top \bs_t} \leq \frac{L^2
  \|\bs_t\|^2}{\by_t^\top \bs_t} \leq \frac{L^2}{\delta},
\label{eq:upper2}
\end{equation}
establishing the upper bound.
For the lower bound, \eqref{eq:safeguard} implies that
\begin{equation}
	\|\by_t\| \geq \delta \|\bs_t\|, \quad \forall t.
\end{equation}
Therefore,
\[
\frac{\by_{t}^\top \bs_{t}}{\by_{t}^\top \by_{t}} \leq
\frac{\|\bs_{t}\|}{\|\by_{t}\|} \leq \frac{1}{\delta}, \quad \forall t.
\]

%

Following \cite{LiuN89a}, $H_t$ can be obtained equivalently by
\begin{align}
H_t^{(0)} &= \gamma_t I,\nonumber\\
H_t^{(k+1)} &= H_t^{(k)} -
\frac{H_t^{(k)}\bs_{t-m(t) + k} \bs_{t-m(t) + k}^\top H_t^{(k)}}{\bs_{t-m(t) + k}^\top
  H_t^{(k)}\bs_{t-m(t) + k}}
+\frac{\by_{t-m(t) + k}\by_{t-m(t) + k}^\top }{\by_{t-m(t) + k}^\top \bs_{t-m(t) + k}}, \;\; 
k=0,\dots, m(t)-1,
\label{eq:inverse}\\
H_t &= H_t^{(m(t))}.
\nonumber
\end{align}
Therefore, we can bound the trace of $H_t^(k)$ and hence $H_t$ through
\eqref{eq:upper2}.
\begin{align}
	\trace\left(H_t^{(k)}\right) \leq \trace\left(H_t^{(0)}\right) +
	\sum_{j=t - m(t) }^{t-m(t) + k} \frac{\by_{j}^\top\by_{j}}{\by_{j}^\top
	\bs_{j}} \leq \gamma_t N + \frac{k L^2}{\delta},
\quad \forall t,
\label{eq:uppB}
\end{align}
where $N$ is the matrix dimension.
According to \cite{ByrNS94a},
the matrix $H^{(k)}_t$ is equivalent to the inverse of
\begin{align}
	B^{(k)}_t &\coloneqq V^\top_{t-m(t) + k}\cdots V^\top_{t-m(t))} B_t^0 V_{t-
	m(t)}\cdots
V_{t-m(t) + k}
+ \rho_{t-m(t) + k} \bs_{t-m(t) + k} \bs_{t-m(t) + k}^\top +\nonumber\\
&\qquad \sum_{j=t-m(t)}^{t-m(t)-1 + k} \rho_{j} V_{t-m(t) + k}^\top\cdots
V_{j+1}^\top \bs_{j} \bs_{j}^\top V_{j+1} \cdots V_{t-m(t) + k},
\label{eq:Bt}
\end{align}
where for $j\geq 0$,
\begin{equation*}
	V_j \coloneqq I - \rho_j \by_j\bs_j^\top,\quad
\rho_j \coloneqq \frac{1}{ \by_j^\top\bs_j},\quad
B_t^0 = \frac{1}{\gamma_t} I.
\end{equation*}
From the form \eqref{eq:Bt}, it is clear that $B_t^{(k)}$ and hence
$H_t$ are all positive-semidefinite because $\gamma_t \geq 0, \rho_j >
0$ for all $j$ and $t$.
Therefore, from positive semidefiniteness,
\eqref{eq:uppB} implies the existence of $c_1 > 0$ such that
\begin{equation*}
	H_t^{(k)}\preceq c_1 I,\quad k = 0,\dotsc,m(t), \quad \forall t.
\end{equation*}
Next, for its lower bound,
from the formulation for \eqref{eq:inverse} in \cite{LiuN89a}, and the
upper bound $\|H_t^{(k)}\| \leq c_1$, we have
\begin{align*}
\det\left(H_t\right)
=\det\left(H_t^{(0)}\right) \prod_{k=t - m(t)}^{t-1}
		\frac{\by_{k}^\top\bs_{k}}{\bs_{k}^\top \bs_{k}}
		\frac{\bs_{k}^\top\bs_{k}}{\bs_{k}^\top H_t^{(k-t + m(t))}
		\bs_{k}}
		\geq \gamma_t^N \left(\frac{\delta}{c_1} \right)^{m(t)}
		\geq M_1.
\end{align*}
for some $M_1>0$.
From that the eigenvalues of $H_t$ are upper-bounded and
nonnegative, and from the
lower bound of the determinant, the eigenvalues of $H_t$ are also
lower-bounded by a positive value $c_2$, completing the proof.

\item By directly expanding $\nabla \hat f$, we have that for
	any $p_1, p_2$,
	\begin{align*}
	\nabla \hat f(p_1) - \nabla \hat f(p_2)
	= \nabla f(x) + H p_1 - \left(\nabla f(x) +
	H p_2\right)
	= H (p_1 - p_2).
	\end{align*}
	Therefore, we have
	\begin{equation*}
		\frac{\left(\nabla \hat f(p_1) - \nabla \hat
			f(p_2)\right)^\top \left( p_1 - p_2 \right)}{\left\|p_1 -
			p_2\right\|^2}
			= \frac{\left\|p_1 - p_2\right\|_H^2}{\left\|p_1 -
			p_2\right\|^2}
			\in \left[c_2, c_1\right]
	\end{equation*}
	for bounding $\psi_i$ for $i > 0$, and the bound for $\psi_0$ is
	directly from the bounds of $\gamma_t$.
	The combined bound is therefore $[\min\{c_2, \delta\}, \max \{c_1,
	L^2/\delta\}]$.
	Next, we show that the final $\psi_i$ is always upper-bounded.
	The right-hand side of \eqref{eq:dk} is equivalent to the
	following:
	\begin{equation}
	\arg \min_\bd \, \hat Q_{\psi_i}\left(\bd\right) \coloneqq \nabla
	\hat f\left(p^{(i)}\right)^\top \bd +
		\frac{\psi_i \left\|\bd\right\|^2}{2}  + \hat \Psi\left(\bd +
		p\right) - \hat \Psi\left(p\right).
		\label{eq:hatQ}
	\end{equation}
	Denote the solution by $\bd$, then we have $p^{(i+1)}
	= p^{(i)} + \bd$.
	Note that we allow $\bd$ to be an approximate solution.
	Because $H$ is upper-bounded by $c_1$, we have that $\nabla \hat
	f$ is $c_1$-Lipschitz continuous.
	Therefore,
	\begin{align}
	Q\left(p^{(i+1)}\right) - Q\left(p^{(i)}\right)
	\nonumber
	\leq&~ \nabla \hat f(p^{(i)})^\top \left( p^{(i+1)} - p^{(i)}
	\right) + \frac{c_1}{2} \left\|p^{(i+1)} - p^{(i)}\right\|^2
	+ \hat \Psi\left(p^{(i+1)}\right) - \hat \Psi\left( p^{(i)}
	\right) \\
	\label{eq:tmp}
	\stackrel{\eqref{eq:hatQ}}{=}&~ \hat Q_{\psi_i}(\bd) -
	\frac{\psi_i}{2} \left\| \bd \right\|^2 + \frac{c_1}{2}
	\left\|\bd\right\|^2.
	\end{align}
	As $\hat Q_{\psi_i}(0) = 0$, provided that the approximate
	solution $\bd$ is better than the point $0$,
	we have
	\begin{equation}
		\hat Q (\bd) \leq \hat Q(0) = 0.
		\label{eq:Qbound}
	\end{equation}
	Putting \eqref{eq:Qbound} into \eqref{eq:tmp}, we obtain
	\begin{equation*}
	Q\left(p^{(i+1)}\right) - Q\left(p^{(i)}\right)
	\leq
\frac{c_1 - \psi_i}{2}\|\bd\|^2.
	\end{equation*}
	Therefore, whenever
	\begin{equation*}
		\frac{c_1 - \psi_i}{2}  \leq -\frac{\sigma_0 \psi_i}{2},
	\end{equation*}
	\eqref{eq:accept} holds.
	This is equivalent to
	\begin{equation*}
		\psi_i \geq \frac{c_1}{1 - \sigma_0},
	\end{equation*}
	Note that the initialization of $\psi_i$ is upper-bounded by
	$c_1$ for all $i > 1$, so the final $\psi_i$ is indeed upper-bounded.
	Together with the first iteration where we start with $\psi_0 =
	\gamma_t$,
	we have that $\psi_i$ for all $i$ are always bounded from the boundedness of
	$\gamma_t$.

	\item
From the results above, at every iteration, SpaRSA finds the update
direction by constructing and optimizing a quadratic
approximation of $\hat f(x)$, where the quadratic term is a multiple
of identity, and its coefficient is bounded in a positive range.
Therefore, the theory developed by \cite{LeeW18a} can be directly used
to show the desired result even if \eqref{eq:dk} is solved only approximately.
For completeness, we provide a simple proof for the case that
\eqref{eq:dk} is solved exactly.

We note that since $Q$ is $c_2$-strongly convex,
the following condition holds.
	\begin{equation}
\frac{\min_{\bs \in \nabla \hat f\left(p^{(i + 1)}\right) + \partial \hat
g\left( p^{(i+ 1)} \right)}
	\left\|\bs \right\|^2}{2 c_2} \geq Q\left(p^{(i + 1)}\right) -
	Q^*.
	\label{eq:kl}
	\end{equation}
	On the other hand, from the optimality condition of
	\eqref{eq:hatQ}, we have that for the optimal solution $\bd^*$ of
	\eqref{eq:hatQ},
	\begin{equation}
		-\psi_i \bd^* = \nabla \hat f\left(p^{(i)}\right) + \bs_{i+1},
		\label{eq:dopt}
	\end{equation}
	for some
	\begin{equation*}
	\bs_{i+1} \in \partial \hat \Psi\left( p^{(i+1)}\right).
	\end{equation*}
	Therefore,
	\begin{align}
		\nonumber
		Q\left( p^{(i+1)} \right) - Q^*
		\stackrel{\eqref{eq:kl}}{\leq}&~ \frac{1}{2 c_2}
		\left\|\nabla \hat f\left( p^{(i+1)} \right) - \nabla \hat
		f\left( p^{(i)} \right) + \nabla \hat f\left( p^{(i)}
		\right) + \bs_{i+1}\right\|^2\\
		\nonumber
		\stackrel{\eqref{eq:dopt}}{\leq}&~ \frac{1}{c_2} \left\|
			\nabla \hat f\left( p^{(i+1)} \right)
		- \nabla \hat f\left( p^{(i)} \right) \right\|^2 + \left\|\psi_i
		\bd^*\right\|^2\\
		\label{eq:dbound}
		\leq&~ \frac{1}{c_2} \left( c_1^2 + \psi_i^2 \right)
		\left\|\bd^*\right\|^2.
	\end{align}
	By combining \eqref{eq:accept} and \eqref{eq:dbound}, we obtain
	\begin{align*}
		Q\left( p^{(i+1)} \right) - Q\left( p^{(i)} \right)
		\leq -\frac{\sigma_0 \psi_i}{2}\left\|\bd^*\right\|^2
		\leq -\frac{\sigma_0 \psi_i}{2} \frac{c_2}{c_1^2 + \psi_i^2}
		\left(Q\left( p^{(i+1)} \right) - Q^*\right).
	\end{align*}
	Rearranging the terms, we obtain
	\begin{equation*}
		\left(1 + \frac{c_2\sigma_0 \psi_i}{2 ( c_1^2 +
		\psi^2)}\right) \left(Q\left( p^{(i+1)} \right) - Q^*\right)
		\leq Q\left( p^{(i)} \right) - Q^*,
	\end{equation*}
	showing Q-linear convergence of SpaRSA,
	with
	\begin{equation*}
		\eta = \sup_{i=0,1,\dotsc}\quad \left( 1 + \frac{c_2 \sigma_0 \psi_i}{2 \left( c_1^2
			+ \psi_i^2
		\right)} \right)^{-1} \in [0,1).
	\end{equation*}
	Note that since $\psi_i$ are bounded in a positive range, we can
	find this supremum in the desired range.
\end{compactenum}
\end{proof}


\section{Implementation Details and Parameter Selection for the Catalyst Framework}
\label{app:catalyst}
We first give an overview to the version of Catalyst
framework for strongly-convex problems \citep{LinMH15a} for
accelerating convergence rate of first-order methods,
then describe our implementation details in the experiment in Section
\ref{subsec:dualexp}.
The Catalyst framework is described in Algorithm \ref{alg:catalyst}.

\begin{algorithm}
\label{alg:catalyst}
\caption{Catalyst Framework for optimizing strongly-convex
	\eqref{eq:f}. }
\begin{algorithmic}[1]
\STATE Input: $x^0 \in \R^N$, a smoothing parameter $\kappa$, the
strong convexity parameter $\mu$, an optimization method $\M$, and a
stopping criterion for the inner optimization.
\STATE Initialize $y^0 = x^0$, $q = \mu / (\mu + \kappa)$, $\beta = (1
- \sqrt{q}) / (1 + \sqrt{q})$.
\FOR{$k=1,2,\dotsc,$}
\STATE Use $\M$ with the input stopping condition to approximately
optimize
\begin{equation}
	\min_x\quad F(x) + \frac{\kappa}{2}\|x - y^{k-1}\|^2
	\label{eq:catalyst}
\end{equation}
from a warm-start point $x^k_0$ to obtain the iterate $x^k$.
\STATE $y_k = x^k + \beta (x^k - x^{k-1})$.
\ENDFOR
\STATE Output $x^k$.
\end{algorithmic}
\end{algorithm}

According to \cite{LinMH15a}, when $\M$ is the proximal gradient
method, the ideal value of $\kappa$ is $\max(L - 2\mu, 0)$, and when
$L > 2\mu$, the convergence speed can be improved to the same order as
accelerated proximal gradient (up to a logarithm factor difference).
Similarly, when $\M$ is stochastic proximal coordinate descent with
uniform sampling, by taking $\kappa = \max(L_{\max} - 2\mu, 0)$,
where $L_{\max}$ is the largest block Lipschitz constant,
one can obtain convergence rate similar to that of accelerated
coordinate descent.
Since when using proximal coordinate descent as the local solver, both
BDA and ADN interpolate between proximal coordinate descent and
proximal gradient,\footnote{Although we used RPCD but not stochastic
	coordinate descent, namely sampling with replacement, it is
	commonly considered that RPCD behaves similar to, and usually
outperforms slightly, the variant that samples without replacement;
see, for example, analyses in
\cite{LeeW16a,WriL17a} and experiment in \cite{ShaZ13a}.}
depending on the number of machines, it is intuitive that acceleration
should work for them.

Considering \eqref{eq:l2svm}, the problem is clearly strongly convex
with parameter $1 / (2C)$, thus we take $\mu = 1 / (2C)$.
For the stopping condition, we use the simple fixed iteration choice
suggested in \cite{LinMH15a} (called (C3) in their notation).
Empirically we found a very effective way is to run $K$ iterations of
BDA with early termination whenever a negative step size is obtained
from exact line search.
For the warm-start part, although \eqref{eq:l2svm} is a regularized
problem, the objective part is smooth, so we take their
suggestion for smooth problem to use $x^k_0 = x^{k-1}$.
Note that they suggested that for general regularized problems,
one should take one proximal gradient step of the original $F$ at
$x^{k-1}$ to obtain $x^k_0$.
We also experimented with this choice, but preliminary results show
that using $x^{k-1}$ gives better initial objective value for
\eqref{eq:catalyst}.

The next problem is how to select $\kappa$.
We observe that for webspam and epsilon, the convergence of both BDA
and ADN clearly falls into two stages.
Through some checks, we found that the first stage can barely be
improved.
On the other hand, if we pick a value of $\kappa$ that can accelerate
convergence at the later stage, the fast early convergence behavior is
not present anymore, thus it takes a long time for the accelerated
approach to outperform the unaccelerated version.
To get better results, we take an approach from the hindsight:
first start with the unaccelerated version with a suitable number of
iterations, and then we switch to Catalyst with $\kappa$ properly
chosen by grid search for accelerating convergence at the later stage.
The parameters in this approach is recorded in Table
\ref{tbl:catalyst}.
We note that this way of tuning from the hindsight favors the
accelerated method unfairly, as it takes information obtained through
running other methods first.
In particular, it requires the optimal objective (obtained by first
solving the problem through other methods) and running the
unaccelerated method to know the turning point of the convergence
stages (requires the optimal objective to compute).
Parameter tuning for $\kappa$ is also needed.
These additional efforts are not included in the running time
comparison, so our experimental result does not suggest that the
accelerated method is better than the unaccelerated version.
The main purpose is to show that our proposed approach also
outperforms acceleration methods with careful parameter choices.

\begin{table}[tb]
\caption{Catalyst parameters.}
\label{tbl:catalyst}
\centering
\begin{tabular}{@{}l|r|r}
Data set & \#BDA iterations before starting Catalyst & $\kappa$\\
\hline
news & $0$ & $17$\\
epsilon & $2,000$ & $12,000$\\
webspam & $400$ & $2,000$
\end{tabular}
\end{table}
\end{document}